\documentclass[10pt,twocolumn,letterpaper]{article}

\usepackage[pagenumbers]{cvpr} % To force page numbers, e.g. for an arXiv version
\usepackage[accsupp]{axessibility} % Improves PDF readability for those with disabilities.
\usepackage{algorithm}
\usepackage{algorithmic}
\usepackage{graphicx}
\usepackage{cite}
\usepackage{caption}
\usepackage{epigraph}
\usepackage[switch]{lineno}
\usepackage{mathtools}
\usepackage{amssymb} 
\usepackage{pifont}
\usepackage{multirow}
\usepackage{multicol}
\usepackage{amsmath,amssymb,amsfonts}
\usepackage{graphicx}
\usepackage[table,xcdraw]{xcolor}
\usepackage{url}
\usepackage{amsmath}

\usepackage{diagbox}
\usepackage{hyperref}
\usepackage{url}
\usepackage{rotating}

\usepackage{amsthm}
\usepackage{bm}
\newtheorem{theorem}{Theorem}

\usepackage{rotating}
\definecolor{cvprblue}{rgb}{0.21,0.49,0.74}
\def\paperID{16924} % *** Enter the Paper ID here
\def\confName{CVPR}
\def\confYear{2025}

\title{Fine-Grained Erasure in Text-to-Image Diffusion-based Foundation Models}

\author{Kartik Thakral$^1$, Tamar Glaser$^2$, Tal Hassner$^3$, Mayank Vatsa$^1$, Richa Singh$^1$ \\
$^1$IIT Jodhpur, $^2$Harman International, $^3$Weir AI \\
{\tt\small \{thakral.1, mvatsa, richa\}@iitj.ac.in}, \tt\small \{tamarglasr, talhassner\}@gmail.com
}
\begin{document}
% \maketitle

 \twocolumn[{
 \maketitle
 \begin{center}
     \captionsetup{type=figure}    \includegraphics[width=0.9\textwidth]{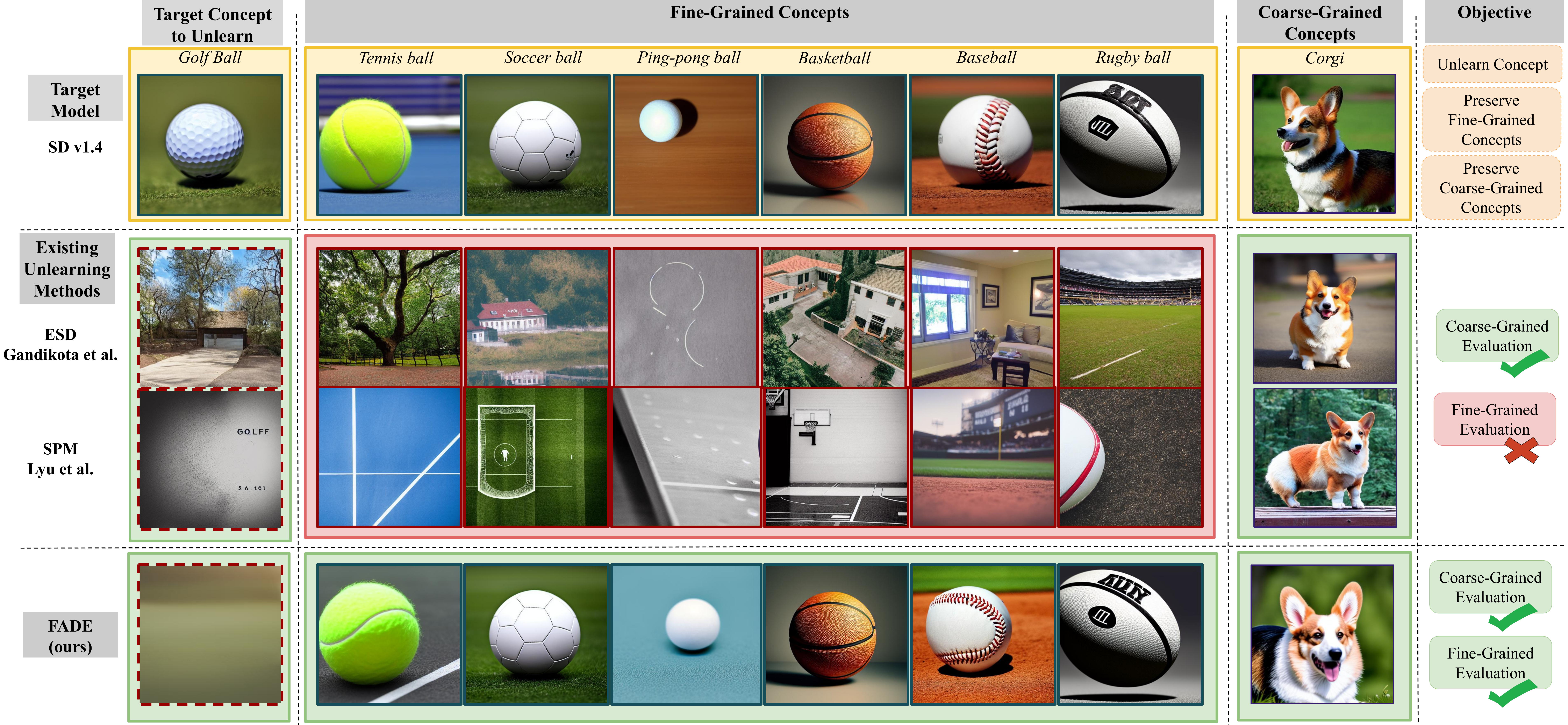}
     \vspace{-4pt}
     \captionof{figure}{Fine-Grained Concept Erasure: This figure demonstrates the issue of collateral forgetting (termed as \textit{adjacency}) in selective concept erasure using existing state-of-the-art algorithms in text-to-image diffusion-based foundation models. It highlights the inability of methods that can precisely erase target concepts from a model’s knowledge while preserving its ability to generate closely related concepts.}
     \label{fig:viz_abs}
\end{center}
 }]

\begin{abstract}

\vspace{-10pt}
Existing unlearning algorithms in text-to-image generative models often fail to preserve the knowledge of semantically related concepts when removing specific target concepts—a challenge known as \textit{adjacency}. To address this, we propose \textbf{FADE} (\textit{Fine-grained Attenuation for Diffusion Erasure}), introducing adjacency-aware unlearning in diffusion models. FADE comprises two components: (1) the \textbf{Concept Neighborhood}, which identifies an adjacency set of related concepts, and (2) \textbf{Mesh Modules}, employing a structured combination of Expungement, Adjacency, and Guidance loss components. These enable precise erasure of target concepts while preserving fidelity across related and unrelated concepts. Evaluated on datasets like Stanford Dogs, Oxford Flowers, CUB, I2P, Imagenette, and ImageNet-1k, FADE effectively removes target concepts with minimal impact on correlated concepts, achieving at least a \textbf{12\% improvement in retention performance} over state-of-the-art methods.  Our code and models are available on the project page: \href{https://www.iab-rubric.org/unlearning-fine-grained-erasure}{iab-rubric/unlearning/FG-Un}.

%This demonstrates FADE's superior control and reduced collateral forgetting, setting a new standard in adjacency-aware unlearning.

%Comparative evaluations highlight FADE’s superior control over unlearning and reduced collateral forgetting, advancing precise, adjacency-aware unlearning in generative models.

%The ABSTRACT is to be in fully justified italicized text, at the top of the left-hand column, below the author and affiliation information. Use the word ``Abstract'' as the title, in 12-point Times, boldface type, centered relative to the column, initially capitalized. The abstract is to be in 10-point, single-spaced type. Leave two blank lines after the Abstract, then begin the main text. Look at previous \confName abstracts to get a feel for style and length.
\end{abstract}    

\vspace{-12pt}
\section{Introduction}
\vspace{-4pt}
\label{sec:intro}
Text-to-image diffusion models \cite{nichol2021glide, ramesh2022hierarchical, rombach2022high} have achieved remarkable success in high-fidelity image generation, demonstrating adaptability across both creative and industrial applications. Trained on expansive datasets like LAION-5B \cite{schuhmann2022laion}, these models capture a broad spectrum of concepts, encompassing diverse objects, styles, and scenes. However, their comprehensive training introduces ethical and regulatory challenges, as these models often retain detailed representations of sensitive or inappropriate content. Thus, there is a growing need for selective concept erasure that avoids extensive retraining, as retraining remains computationally prohibitive \cite{esd, CA, sinitsin2020editable}.

Current generative unlearning methods aim to remove specific concepts while preserving generation capabilities for unrelated classes, focusing on the concept of \textbf{locality} \cite{esd, CA, SPM, receler}. However, these methods often lack fine-grained control, inadvertently affecting semantically similar classes when erasing a target concept (refer Figure \ref{fig:viz_abs}). This creates the need for \textbf{adjacency-aware unlearning}—the ability to retain knowledge of classes closely related to the erased concept. Specifically, adjacency-aware unlearning seeks to modify a model such that the probability of generating the target concept \( c_{\text{tar}} \) given input \( x \) approaches zero, i.e., \( P_\theta(c_{\text{tar}} | x) \to 0 \), while ensuring \( P_\theta(\mathcal{A}(c_{\text{tar}}) | x) \approx P_{\theta_{\text{original}}}(\mathcal{A}(c_{\text{tar}}) | x) \), where \( \mathcal{A}(c_{\text{tar}}) \) represents a carefully constructed set of semantically related classes that should remain unaffected by unlearning.

%To address these challenges, we introduce \textbf{FADE} (\textit{Fine-grained Attenuation for Diffusion Erasure}), a framework specifically designed for adjacency-aware unlearning in text-to-image diffusion models. FADE comprises two core components: the \textbf{Concept Lattice}, which systematically identifies semantically related classes to construct an adjacency set by leveraging fine-grained semantic similarity for targeted unlearning, and the \textbf{Mesh Modules}, which utilize a structured combination of Expungement, Adjacency, and Guidance loss components to balance the erasure of target concepts with the retention of adjacent classes, thus preserving the semantic integrity of neighboring and unrelated concepts.

To address these challenges, we introduce \textbf{FADE} (\textit{Fine-grained Attenuation for Diffusion Erasure}), a framework for adjacency-aware unlearning in text-to-image diffusion models. FADE has two core components: the \textbf{Concept Neighborhood}, which identifies semantically related classes to form an adjacency set using fine-grained semantic similarity, and the \textbf{Mesh Modules}, which balance target concept erasure with adjacent class retention through Expungement, Adjacency, and Guidance loss components. This design ensures effective unlearning of target concepts while preserving the integrity of neighboring and unrelated concepts in the semantic manifold. We evaluate FADE using the Erasing-Retention Balance Score (ERB), the proposed metric that quantifies both forgetting and adjacency retention. Experimental results across fine- and coarse-grained datasets—including Stanford Dogs, Oxford Flowers, CUB, I2P, Imagenette, and ImageNet-1k—demonstrate FADE's effectiveness in erasing targeted concepts while protecting representations of adjacent classes. The key contributions include (i) formalization of adjacency-aware unlearning for text-to-image diffusion models, emphasizing the need for precise retention control, (ii) introduction of FADE, a novel method for unlearning target concepts with effective adjacency retention, and (iii) proposal of the Erasing-Retention Balance Score (ERB) metric, designed to capture both forgetting efficacy and adjacency retention. Using ERB, extensive evaluations are performed on fine- and coarse-grained protocols to assess the erasing performance of FADE compared to state-of-the-art methods.
%Comparative analysis with existing unlearning methods highlights FADE’s superior control and significantly reduced collateral forgetting.

% \vspace{-3pt}
\section{Related Work}
% \vspace{-3pt}
Advancements in generative modeling have highlighted the need for effective unlearning techniques. Text-to-image generative models trained on large datasets often encapsulate undesired or inappropriate content, necessitating methods that can selectively remove targeted concepts while preserving overall model functionality. Generative machine unlearning aims to facilitate precise modifications without affecting unrelated knowledge.

\vspace{1mm}
\noindent\textbf{Generative Machine Unlearning:} Existing approaches focus on unlearning specific concepts from generative models. Gandikota et al.\ \cite{esd} used negative guidance in diffusion models to steer the generation away from unwanted visual elements like styles or object classes. FMN \cite{FMN} adjusts cross-attention mechanisms to reduce emphasis on undesired concepts, while Kumari et al.\ \cite{CA} aligned target concepts with surrogate embeddings to guide models away from undesirable outputs. Thakral et al. \cite{DUGE} proposed a robust method of continual unlearning for sequential erasure of concepts. For GANs, Tiwari et al.\ \cite{tiwary2023adapt} introduced adaptive retraining to selectively erase classes. However, the high computational cost of retraining remains a drawback \cite{esd, CA, sinitsin2020editable}, highlighting the need for efficient methods.

Parameter-Efficient Fine-Tuning (PEFT) methods address these computational challenges by modifying a small subset of parameters. UCE \cite{gandikota2024unified} offers a closed-form editing approach that aligns target embeddings with surrogates, enabling concept erasure while preserving unrelated knowledge. SPM \cite{SPM} introduced "Membranes," lightweight adapters that selectively erase concepts by altering model sensitivity. Similarly, Receler \cite{receler} incorporates "Erasers" into diffusion models, for robust and adversarially resilient concept erasure with minimal impact on unrelated~content.

\vspace{1mm}
\noindent\textbf{Fine-Grained Classification:}
Fine-grained classification tackles the challenge of distinguishing highly similar classes, often complicated by subtle visual differences and label ambiguities. In datasets like ImageNet \cite{imagenet_1k_dataset}, overlapping characteristics between classes hinder classification accuracy. Beyer et al.\ \cite{beyer2020we} and Shankar et al.\ \cite{shankar2020evaluating} introduced multi-label evaluation protocols to accommodate multiple entities within a single image, benefiting tasks such as organism classification.

Recent methods have advanced the evaluation of fine-grained errors by automating their categorization. Vasudevan et al.\ \cite{vasudevan2022does} proposed an error taxonomy to distinguish fine-grained misclassifications from out-of-vocabulary (OOV) errors, enabling nuanced analyses of visually similar classes. Peychev et al.\ \cite{peychev2023automated} automated error classification, providing deeper insights into model behavior in fine-grained scenarios.

\vspace{1mm}

\noindent\textbf{Challenges in Adjacency-Aware Erasure:}
Despite progress, achieving adjacency-aware erasure while maintaining locality remains a significant challenge. Current unlearning methods often struggle with fine-grained concept forgetting, inadvertently affecting the semantic neighborhood of the target concept. This underscores the need for techniques that can selectively remove only the target concept while preserving the integrity of related classes.

% \begin{figure*}[t]
% \centering
%   \includegraphics[width=0.9\textwidth]{Images/FADE_M.pdf}
% \caption{Visual illustration of the complete erasure process. (a) The dataset $D$ is organized into an unlearning set $\mathcal{D}_u$ and an adjacency set $\mathcal{A}(c_{\text{tar}})$ using the concept lattice. (b) These sets are utilized by mesh modules for selective erasure while maintaining semantic integrity on neighboring concepts.} 
%     \label{fig:gen_pipeline}
% \end{figure*}

\begin{figure*}[t]
\centering
  \includegraphics[width=\textwidth]{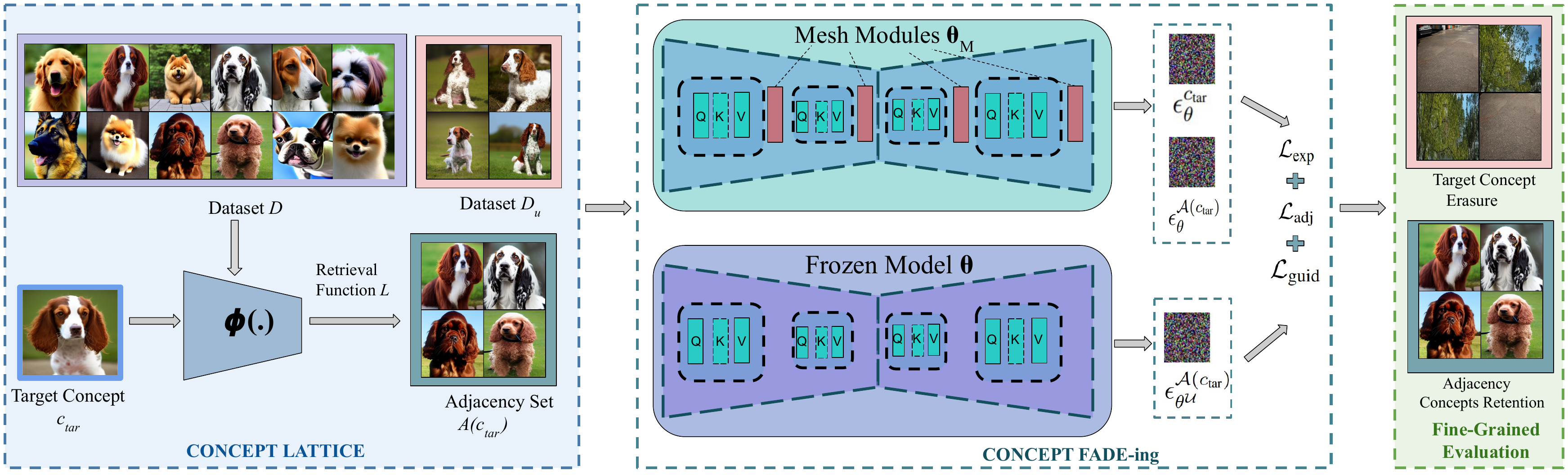}
\caption{Visual illustration of complete erasure process. (a) The dataset $D$ is organized into unlearning set $\mathcal{D}_u$ and adjacency set $\mathcal{A}(c_{\text{tar}})$ using concept neighborhood, (b) these sets are utilized by mesh-modules for selective erasure while maintaining semantic integrity if the model on neighboring concepts.} 
    \label{fig:gen_pipeline}
\end{figure*}

%-------------------------------------------------------------------------
% \vspace{-2pt}
\section{Fine-Grained Unlearning}
% \vspace{-2pt}
\subsection{Preliminary}
% \vspace{-2pt}
Text-to-image diffusion models have become essential for high-quality image synthesis by learning to generate images through a denoising process \cite{ho2020denoising, SDM}. Starting with Gaussian noise, these models refine an image over $T$ timesteps by predicting the noise component $\epsilon_\theta(x_t, c, t)$ at each step $t$, conditioned on a textual prompt $c$. This reverse process, modeled as a Markov chain, aims to recover the final image $x_0$ from initial noise $x_T$, with generation probability defined as $P_\theta(x_{0:T}) = P(x_T) \prod_{t=1}^T P_\theta(x_{t-1}|x_t)$, where $P(x_T)$ is the Gaussian prior. Latent Diffusion Models (LDMs) \cite{rombach2022high} further improve efficiency by operating in a compressed latent space $z$, where $z = \mathcal{E}(x)$, and noise is progressively added to obtain $z_t$. The model learns to minimize the difference between true noise $\epsilon$ and predicted noise $\epsilon_\theta$, with a training objective:

% \vspace{-5pt}
\begin{equation}
\mathcal{L} = \mathbb{E}_{z, t, c, \epsilon} \left[ \left| \epsilon - \epsilon{_\theta}(z_t, c, t) \right|_2^2 \right], \; \epsilon \sim \mathcal{N}(0,1),
\end{equation}

Given the high parameter count of these models, efficient fine-tuning for unlearning tasks necessitates Parameter-Efficient Fine-Tuning (PEFT) techniques. We employ a LoRA-based method~\cite{hu2021lora}, termed \textbf{Mesh Modules} throughout this paper, which selectively updates only a subset of model parameters. Specifically, the weight update $\nabla W$ for any pretrained weight matrix $W_0 \in \mathbb{R}^{d \times k}$ is decomposed as $\nabla W = B A$, where $B \in \mathbb{R}^{d \times r}$, $A \in \mathbb{R}^{r \times k}$, and $r \ll \min(d, k)$. Only the smaller matrices $A$ and $B$ are trained, preserving computational efficiency and limiting the risk of overfitting by keeping $W_0$ fixed. This adaptation effectively enables precise concept removal while preserving core generative capabilities.

% By leveraging Mesh modules, our approach enables precise unlearning of the target concept  $c_{\text{tar}}$, enhancing computational efficiency while preserving model performance on both neighboring and unrelated concepts.

% Maybe combine prilimaries into a single section.

\subsection{Problem Formulation}
% \vspace{-3pt}
\label{sec:problem_definition}
Our objective is to selectively unlearn a target concept  $c_{\text{tar}} \in \mathcal{C}$  in a generative model while preserving its performance on semantically similar (adjacent) and unrelated concepts. Let $\mathcal{D} = \{d_1, d_2, \dots, d_N\}$ represent a dataset where each data point  $d_i$  is associated with a subset of concepts $\mathcal{C}_{d_i} \subseteq \mathcal{C}$, with $\mathcal{C}$ representing the universal set of all concepts learned by the model. We denote the pre-trained generative model as $\theta$, mapping input prompts $x \in \mathcal{X}$ to images $y \in \mathcal{Y}$, thereby learning the conditional distribution  $P_{\theta}(y \mid x)$. To achieve unlearning, we aim to update the model parameters from $\theta$ to $\theta^{\mathcal{U}}$ via an unlearning function $\mathcal{U}$, such that the probability of generating images associated with $c_{\text{tar}}$ i.e., $y_{c_{\text{tar}}}$ approaches zero for any input prompt $x$, expressed as $P_{\theta^{\mathcal{U}}}(y_{c_{\text{tar}}}\mid x) \rightarrow 0, \; \forall x \in \mathcal{X}$. Simultaneously, we seek to maintain the model’s performance on adjacent concepts and unrelated concepts.

Let $\mathcal{A}(c_{\text{tar}}) \subseteq \mathcal{C}$ denote the adjacency set, containing concepts closely related to $c_{\text{tar}}$. The unlearning objective must satisfy the following:

\begin{enumerate}
    \item Retention of Adjacent Concepts:
% \vspace{-5pt}
\begin{equation}
P_{\theta^{\mathcal{U}}}(y_c \mid x) \approx P_{\theta}(y_c \mid x), \, \forall c \in \mathcal{A}(c_{\text{tar}}), \forall x \in \mathcal{X}.
\end{equation}

    \item Preservation of Unrelated Concepts:
% \vspace{-5pt}
\begin{equation}
P_{\theta^{\mathcal{U}}}(y_c \mid x) \approx P_{\theta}(y_c \mid x), \, \forall c \in \mathcal{C} \setminus { c_{\text{tar}} \cup \mathcal{A}(c_{\text{tar}}) }, \forall x \in \mathcal{X}.
\end{equation}
\end{enumerate}

% To measure the impact of unlearning, we define $\Delta \mathcal{L}(c) = \mathcal{L}{\theta^{\mathcal{U}}}(c) - \mathcal{L}{\theta}(c)$, where $\mathcal{L}_{M}(c)$ denotes an appropriate evaluation metric (e.g., generation quality, classification accuracy) for each concept $c$. Two concepts $c_i$ and $c_j$ are considered correlated if unlearning $c_i$ leads to a significant change in performance on $c_j$, i.e., if  $|\Delta \mathcal{L}(c_j)| > \delta$  for a threshold $\delta > 0$. Our goal is to minimize  $|\Delta \mathcal{L}(c)|$  for all  $c \in \mathcal{A}(c_{\text{tar}}) \cup \mathcal{C} \setminus \{ c_{\text{tar}} \}$ , ensuring that unlearning the target concept  $c_{\text{tar}}$  does not adversely affect the model’s ability to generate images for neighboring or distant concepts.

% FineFade, AuraMesh
% FADE: Fine-grained Attenuation for Diffusion Erasure
% FADE: Fine-grAined Diffusion Erasure
%\vspace{-10pt}

% \vspace{-6pt}
\subsection{FADE: Fine-grained Erasure}
% \vspace{-2pt}
We present FADE (Fine-grained Attenuation for Diffusion Erasure), a method for targeted unlearning in text-to-image generative models, designed to remove specific concepts while preserving fidelity on adjacent and unrelated concepts (see Figure \ref{fig:viz_abs}). FADE organizes model knowledge into three subsets: the Unlearning Set $\mathcal{D}_u$, the Adjacency Set $\mathcal{D}_a$, and the Retain Set $\mathcal{D}_r$.

The Unlearning Set $\mathcal{D}_u$ consists of images generated using the target concept $c_{\text{tar}}$, such as “Golden Retriever” for a retriever breed class. The Adjacency Set $\mathcal{D}_a$ contains images of concepts similar to $c_{\text{tar}}$ (e.g., related retriever breeds), ensuring the erasure of $c_{\text{tar}}$ does not compromise the model’s ability to generate closely related classes. We construct $\mathcal{D}_a$ using Concept Neighborhood, which systematically identifies semantically proximal classes to $c_{\text{tar}}$ based on similarity scores.

The Retain Set $\mathcal{D}_r$, containing images of diverse and unrelated concepts (e.g., “Cat” or “Car”), serves as a check for broader generalization retention. While successful retention on $\mathcal{D}_a$ typically implies generalization to $\mathcal{D}_r$, testing with $\mathcal{D}_r$ ensures no unintended degradation in unrelated areas.

FADE employs a structured mesh to modulate the likelihood of generating images including $c_{\text{tar}}$, gradually attenuating the concept’s influence while preserving related and unrelated knowledge. We formalize this data organization by ensuring $\mathcal{D}_u \cup \mathcal{D}_a \cup \mathcal{D}_r \subseteq \mathcal{D}$, with $\mathcal{D}_u \cap \mathcal{D}_a \cap \mathcal{D}_r = \emptyset$. The complete framework can be visualized in Figure \ref{fig:gen_pipeline}.

% \subsection{Estimating Adjacency Set via Concept Lattice}
\vspace{1mm}
\noindent\textbf{Concept Neighborhood - Synthesizing Adjacency Set}
Evaluating unlearning on fine-grained concepts requires an adjacency set $\mathcal{D}_a$, designed to preserve the model’s performance on concepts neighboring the target concept $c_{\text{tar}}$. Ideally, $\mathcal{D}_a = \{ c \in \mathcal{C} \mid \text{sim}(c, c_{\text{tar}}) > \tau \}$, where $\text{sim}(c, c_{\text{tar}})$ represents a semantic similarity function and $\tau$ is a threshold for high similarity. However, in the absence of taxonomical hierarchies or semantic annotations (e.g., WordNet synsets), constructing $\mathcal{D}_a$ becomes challenging. To address this, we propose an approximation $\mathcal{A}(c_{\text{tar}})$, termed the \textit{Concept Neighborhood}, which leverages semantic similarities to identify the top-K classes most similar to $c_{\text{tar}}$ and thus serves as a practical substitute for $\mathcal{D}_a$.

To construct $\mathcal{A}(c_{\text{tar}})$, we proceed as follows: for each concept $c \in \mathcal{C}$, including $c_{\text{tar}}$, we generate a set of images $\mathcal{I}_{c} = \{ x_{1}^{c}, x_{2}^{c}, \dots, x_{m}^{c} \}$ using $\theta$, where $m$ is the number of images per concept. Using a pre-trained image encoder $\phi: X \rightarrow \mathbb{R}^{d}$, we compute embeddings for each image: $\mathbf{f}_{i}^{c} = \phi(x_{i}^{c})$ for all $x_{i}^{c} \in \mathcal{I}_{c}$. For each concept $c$, we then compute the mean feature vector $\bar{\mathbf{f}}^{c} = \frac{1}{N} \sum_{i=1}^{N} \mathbf{f}_{i}^{c}$ and quantify the semantic similarity between the target concept $c_{\text{tar}}$ and every other concept $c \in \mathcal{C} \setminus \{ c_{\text{tar}} \}$ by calculating the cosine similarity between their mean feature vectors:

% \vspace{-3pt}
\begin{equation}
L(c_{\text{tar}}, c) = \frac{ \left\langle \bar{\mathbf{f}}^{c_{\text{tar}}}, \bar{\mathbf{f}}^{c} \right\rangle }{ \left| \bar{\mathbf{f}}^{c_{\text{tar}}} \right| \left| \bar{\mathbf{f}}^{c} \right| },
\end{equation}

where $\left\langle \cdot, \cdot \right\rangle$ denotes the inner product, and $\left\| \cdot \right\|$ denotes the Euclidean norm. We select the top-$K$ concepts with the highest similarity to $c_{\text{tar}}$ to form the adjacency set $\mathcal{A}(c_{\text{tar}}) = \left\{ c^{(1)}, c^{(2)}, \dots, c^{(K)} \right\}$, where $L(c_{\text{tar}}, c^{(i)}) \geq L(c_{\text{tar}}, c^{(i+1)})$ for $i = 1, \dots, K-1$, and $c^{(i)} \in \mathcal{C} \setminus \{ c_{\text{tar}} \}$.

This approach effectively constructs $\mathcal{A}(c_{\text{tar}})$ by capturing the fine-grained semantic relationships inherent in the latent feature space, approximating the ideal $\mathcal{D}_a$ with a data-driven methodology that leverages embedding similarity.

Our Concept Neighborhood method is further supported by a theoretical link between k-Nearest Neighbors (k-NN) classification in latent feature space and the optimal Naive Bayes classifier under certain conditions, as established in the following theorem:

\iffalse
\noindent\textit{Theorem (k-NN Approximation to Naive Bayes in $\mathbb{R}^d)$.} Let  $\mathbf{x} \in \mathbb{R}^{h \times w \times c}$  represent an image with dimensions $h$,  $w$, and $c$, respectively, and let  $\phi: \mathbb{R}^{h \times w \times c} \to \mathbb{R}^d$  be the mapping function of $\mathbf{x}$ to a latent feature space $\mathbb{R}^d$, where  $d \ll hwc$. Assuming the latent features  $z := \phi(\mathbf{x})$  are conditionally independent given the class label  $C \in \mathcal{C}$, the k-Nearest Neighbors (k-NN) classifier operating in $\mathbb{R}^d$ converges to the Bayes optimal classifier as the sample size $N \to \infty$ and the number of neighbors $k \to \infty$, with $k/N \to 0$, i.e.,

% \vspace{-4pt}
\begin{equation}
\lim_{N \to \infty} P(C_{\text{K-NN}}(\phi(\mathbf{x})) = C_{\text{NB}}(\mathbf{x})) = 1.
\end{equation}
\fi

\vspace{4pt}
\noindent\rule{\linewidth}{0.4pt}
\vspace{-20pt}

\begin{theorem}[k-NN Approximation to Naive Bayes in $\mathbb{R}^d$]
Let $\mathbf{x} \in \mathbb{R}^{h \times w \times c}$ represent an image with dimensions height $h$, width $w$, and channels $c$. Let the mapping function 
$\phi: \mathbb{R}^{h \times w \times c} \to \mathbb{R}^d$ project the image $\mathbf{x}$ into a latent feature space $\mathbb{R}^d$, where $d \ll hwc$. Assume that the latent features $z := \phi(\mathbf{x})$ are conditionally independent given the class label $C \in \mathcal{C}$. 

Then, the k-Nearest Neighbors (k-NN) classifier operating in $\mathbb{R}^d$ converges to the Naive Bayes classifier as the sample size $N \to \infty$, the number of neighbors $k \to \infty$, and $k/N \to 0$. Specifically,
\begin{equation}
\lim_{N \to \infty} P\bigl(C_{\text{k-NN}}(\phi(\mathbf{x})) = C_{\text{NB}}(\mathbf{x})\bigr) = 1.
\end{equation}
\end{theorem}
\vspace{-8pt}
\noindent\rule{\linewidth}{0.4pt}
% \vspace{-12pt}

A detailed proof is available in the supplementary material, but intuitively, this result shows that the k-NN classifier in latent space approximates the optimal classifier, supporting the use of feature similarity (via k-NN) to identify semantically similar concepts. Thus, the Concept Neighborhood method approximates the latent space’s underlying semantic structure, effectively constructing $\mathcal{A}(c_{\text{tar}})$ for adjacency preservation in the unlearning framework.

\vspace{1mm}
\noindent\textbf{Concept FADE-ing}
The proposed FADE (Fine-grained Attenuation for Diffusion Erasure) algorithm selectively unlearns a target concept $c_{\text{tar}}$ through the mesh $M$, parameterized by $ \theta^{\mathcal{U}}_M$, while maintaining the model’s semantic integrity for neighboring concepts. FADE achieves this by optimizing three distinct loss terms: the Erasing Loss, the Guidance Loss, and the Adjacency Loss.

\begin{enumerate}
\item \textbf{Erasing Loss ($\mathcal{L}_{\text{er}}$):} This loss is designed to encourage the model to erase $c_{\text{tar}}$ by modulating the predicted noise $\epsilon_{\theta}$ in such a way that the changes in the unlearned model are minimal with respect to semantically related classes in $\mathcal{A}(c_{\text{tar}})$, thereby acting as a regularization term. Concurrently, it drives the model’s representation of the target concept $c_{\text{tar}}$ in $\mathcal{D}_u$, disorienting it away from its initial position. Formally, the Erasing Loss is defined as:
% \vspace{-5pt}
\begin{equation}
\begin{split}
% \epsilon_{\theta^{\mathcal{U}}}^{\mathcal{A}(c_{\text{tar}})}}
\mathcal{L}_{\text{er}} = \max\Bigg(0, &\frac{1}{|\mathcal{A}(c_{\text{tar}})|} \sum_{x \in \mathcal{A}(c_{\text{tar}})} \left| \epsilon_{\theta^{\mathcal{U}}_M}^{c_{\text{tar}}} - \epsilon_{\theta}^{x} \right|_2^2 \\
&- \frac{1}{|\mathcal{D}_u|} \sum_{x \in \mathcal{D}_u} \left| \epsilon_{\theta^{\mathcal{U}}_M}^{c_{\text{tar}}} - \epsilon_{\theta}^{x} \right|_2^2 + \delta \Bigg)
\end{split}
\end{equation}

where $\epsilon_{\theta}^{c_{\text{tar}}}$ represents the predicted noise for the target concept, $\epsilon_{\theta}^{x}$ denotes the predicted noise for samples $x$ in either the adjacency set $\mathcal{A}(c_{\text{tar}})$ or the unlearning set $\mathcal{D}_u$, and $\delta$ is a margin hyperparameter enforcing a minimum separation between the noise embeddings of $c_{\text{tar}}$ and its adjacent concepts. 
% This loss enforces a dual objective: pushing $c_{\text{tar}}$ away from related concepts in $\mathcal{A}(c_{\text{tar}})$ to ensure effective erasure, while preserving generalization by keeping $c_{\text{tar}}$ close to unrelated concepts in $\mathcal{D}_r$.

     % Add Guidance scale
    \item \textbf{Guidance Loss ($\mathcal{L}_{\text{guid}}$):} The Guidance Loss directs the noise prediction for $c_{\text{tar}}$ toward a surrogate "null" concept, allowing unlearning without requiring a task-specific surrogate. Formally, it is defined as:

\begin{equation}
\mathcal{L}_{\text{guid}} = \left| \epsilon_{\theta^{\mathcal{U}}_M}^{c_{\text{tar}}} - \epsilon_{\theta}^{c_{\text{null}}} \right|_2^2
\end{equation}

where $\epsilon_{\theta}^{c_{\text{null}}}$ denotes the predicted noise for a neutral or averaged “null” concept in the original model. By directing $c_{\text{tar}}$ toward a null state, this loss effectively nullifies the influence of the target concept, facilitating generalized unlearning that is adaptable across tasks without specific surrogate selection \cite{CA, SPM}.
    
    \item \textbf{Adjacency Loss ($\mathcal{L}_{\text{adj}}$):} The Adjacency Loss acts as a regularization term, preserving the embeddings of concepts in the adjacency set $\mathcal{A}(c_{\text{tar}})$ in the updated model $M_{\theta^{\mathcal{U}}}$. It penalizes deviations between the original and updated model’s noise predictions for these adjacent concepts, defined as:

% \vspace{-5pt}
\begin{equation}
\mathcal{L}_{\text{adj}} = \frac{1}{|\mathcal{A}(c_{\text{tar}})|} \sum_{x \in \mathcal{A}(c_{\text{tar}})} \left| \epsilon_{\theta^{\mathcal{U}}_M}^{x} - \epsilon_{\theta}^{x} \right|_2^2
\end{equation}

where $\epsilon_{\theta}^{x}$ and $\epsilon_{\theta^{\mathcal{U}}}^{x}$ denote the noise predictions for concept x in the original and updated models, respectively. This loss constrains the modified model to retain the structural relationships among adjacent classes, preserving the feature space of $\mathcal{A}(c_{\text{tar}})$ post-unlearning.
\end{enumerate}

\noindent The total loss function for the FADE algorithm is a weighted sum of the three loss terms:

\begin{equation}
\mathcal{L}_{\text{FADE}} = \lambda_{\text{er}} \mathcal{L}_{\text{er}} + \lambda_{\text{adj}} \mathcal{L}_{\text{adj}} + \lambda_{\text{guid}} \mathcal{L}_{\text{guid}}
\end{equation}

where$ \lambda_{\text{er}}$, $\lambda_{\text{adj}}$, and $\lambda_{\text{guid}}$ are hyperparameters controlling the relative influence of each loss term.

% Comparison table on FG-Un on Dogs, Flowers, and CUB dataset.
\begin{table*}[t]
\centering
\resizebox{0.95\linewidth}{!}{%
\begin{tabular}{l|c|ccc|ccc|ccc}
\hline
                                                                                &                                    & \multicolumn{3}{c|}{\textbf{Stanford Dogs}}                                                                                                                        & \multicolumn{3}{c|}{\textbf{Oxford Flowers}}                                                                                                 & \multicolumn{3}{c}{\textbf{CUB}}                                                                                         \\
\multirow{-2}{*}{\textbf{Methods}}                                              & \multirow{-2}{*}{\textbf{Metrics}} & \begin{tabular}[c]{@{}c@{}}Welsh Springer \\ Spaniel\end{tabular} & \begin{tabular}[c]{@{}c@{}}German \\ Shepherd\end{tabular} & Pomeranian                             & \begin{tabular}[c]{@{}c@{}}Barbeton \\ Daisy\end{tabular} & Yellow Iris                            & Blanket Flower                         & Blue Jay                               & Black Tern                             & Barn Swallow                           \\ \hline
                                                                                & $A_{\text{er}} $                    & 100.00                                                            & 100.00                                                     & 100.00                                 & 100.00                                                    & 100.00                                 & 100.00                                 & 100.00                                 & 100.00                                 & 100.00                                 \\
                                                                                & $\hat{A}_{\text{adj}}$                   & 20.00                                                             & 20.00                                                      & 34.00                                  & 48.00                                                     & 6.00                                   & 4.00                                   & 15.00                                  & 8.00                                   & 38.00                                  \\
\multirow{-3}{*}{\begin{tabular}[c]{@{}l@{}}ESD \cite{esd}\\ (ICCV 2023)\end{tabular}}     & \cellcolor[HTML]{CBCEFB}ERB        & \cellcolor[HTML]{CBCEFB}33.34                                     & \cellcolor[HTML]{CBCEFB}33.34                              & \cellcolor[HTML]{CBCEFB}50.74          & \cellcolor[HTML]{CBCEFB}64.86                             & \cellcolor[HTML]{CBCEFB}11.32          & \cellcolor[HTML]{CBCEFB}7.69           & \cellcolor[HTML]{CBCEFB}26.08          & \cellcolor[HTML]{CBCEFB}14.81          & \cellcolor[HTML]{CBCEFB}55.07          \\ \hline
                                                                                & $A_{\text{er}} $                    & 98.80                                                             & 100.00                                                     & 98.20                                  & 75.60                                                     & 100.00                                 & 63.00                                  & 100.00                                 & 96.00                                  & 98.80                                  \\
                                                                                & $\hat{A}_{\text{adj}}$                   & 0.20                                                              & 0.57                                                       & 0.60                                   & 1.96                                                      & 7.44                                   & 0.84                                   & 0.42                                   & 2.76                                   & 4.28                                   \\
\multirow{-3}{*}{\begin{tabular}[c]{@{}l@{}}FMN \cite{FMN}\\ (CVPRw 2024)\end{tabular}}    & \cellcolor[HTML]{CBCEFB}ERB        & \cellcolor[HTML]{CBCEFB}0.39                                      & \cellcolor[HTML]{CBCEFB}1.14                               & \cellcolor[HTML]{CBCEFB}1.19           & \cellcolor[HTML]{CBCEFB}3.82                              & \cellcolor[HTML]{CBCEFB}13.84          & \cellcolor[HTML]{CBCEFB}1.65           & \cellcolor[HTML]{CBCEFB}0.84           & \cellcolor[HTML]{CBCEFB}5.36           & \cellcolor[HTML]{CBCEFB}8.13           \\ \hline
                                                                                & $A_{\text{er}} $                    & 63.00                                                             & 79.20                                                      & 68.00                                  & 70.20                                                     & 67.60                                  & 27.00                                  & 68.60                                  & 77.62                                  & 42.40                                  \\
                                                                                & $\hat{A}_{\text{adj}}$                   & 66.67                                                             & 63.66                                                      & 84.40                                  & 78.57                                                     & 55.36                                  & 78.60                                  & 61.24                                  & 54.04                                  & 77.92                                  \\
\multirow{-3}{*}{\begin{tabular}[c]{@{}l@{}}CA \cite{CA}\\ (ICCV 2023)\end{tabular}}      & \cellcolor[HTML]{CBCEFB}ERB        & \cellcolor[HTML]{CBCEFB}64.75                                     & \cellcolor[HTML]{CBCEFB}70.58                              & \cellcolor[HTML]{CBCEFB}75.31          & \cellcolor[HTML]{CBCEFB}74.15                             & \cellcolor[HTML]{CBCEFB}60.87          & \cellcolor[HTML]{CBCEFB}40.19          & \cellcolor[HTML]{CBCEFB}64.71          & \cellcolor[HTML]{CBCEFB}63.71          & \cellcolor[HTML]{CBCEFB}54.91          \\ \hline
                                                                                & $A_{\text{er}}$   & 98.20  & 100.00  & 100.00  & 99.00  & 98.20  & 99.00 & 100.00 & 100.00 & 100.00     \\
                                                                                & $\hat{A}_{\text{adj}}$             & 41.76                                                             & 46.20                                                      & 50.72                                  & 53.56                                                     & 39.66                                                 & 61.24                                                    & 31.98                                              & 34.88                                                & 43.44                                                                                    \\
\multirow{-3}{*}{\begin{tabular}[c]{@{}l@{}}UCE \cite{gandikota2024unified}\\ (WACV 2024)\end{tabular}}    & \cellcolor[HTML]{CBCEFB}ERB        & \cellcolor[HTML]{CBCEFB}58.80                                     & \cellcolor[HTML]{CBCEFB}63.27                              & \cellcolor[HTML]{CBCEFB}67.30          & \cellcolor[HTML]{CBCEFB}69.51                             & \cellcolor[HTML]{CBCEFB}56.50                         & \cellcolor[HTML]{CBCEFB}75.67                            & \cellcolor[HTML]{CBCEFB}48.46                      & \cellcolor[HTML]{CBCEFB}51.72                        & \cellcolor[HTML]{CBCEFB}60.56                             \\ \hline
                                                                                & $A_{\text{er}} $                    & 57.80                                                             & 99.20                                                      & 33.60                                  & 70.00                                                     & 48.40                                  & 54.00                                  & 85.40                                  & 86.28                                  & 92.60                                  \\
                                                                                & $\hat{A}_{\text{adj}}$                   & 65.12                                                             & 70.80                                                      & 95.20                                  & 91.64                                                     & 81.68                                  & 84.40                                  & 80.24                                  & 62.16                                  & 69.64                                  \\
\multirow{-3}{*}{\begin{tabular}[c]{@{}l@{}}SPM \cite{SPM}\\ (CVPR 2024)\end{tabular}}     & \cellcolor[HTML]{CBCEFB}ERB        & \cellcolor[HTML]{CBCEFB}61.24                                     & \cellcolor[HTML]{CBCEFB}82.62                              & \cellcolor[HTML]{CBCEFB}49.66          & \cellcolor[HTML]{CBCEFB}79.37                             & \cellcolor[HTML]{CBCEFB}60.78          & \cellcolor[HTML]{CBCEFB}65.86          & \cellcolor[HTML]{CBCEFB}82.73          & \cellcolor[HTML]{CBCEFB}72.23          & \cellcolor[HTML]{CBCEFB}79.49          \\ \hline
                                                                                & $A_{\text{er}} $                    & 100.00                                                            & 100.00                                                     & 100.00                                 & 100.00                                                    & 100.00                                 & 100.00                                 & 100.00                                 & 100.00                                 & 100.00                                 \\
                                                                                & $\hat{A}_{\text{adj}}$                   & 2.40                                                              & 2.80                                                       & 1.16                                   & 6.32                                                      & 0.52                                   & 0.88                                   & 0.68                                   & 0.12                                   & 1.28                                   \\
\multirow{-3}{*}{\begin{tabular}[c]{@{}l@{}}Receler \cite{receler}\\ (ECCV 2024)\end{tabular}} & \cellcolor[HTML]{CBCEFB}ERB        & \cellcolor[HTML]{CBCEFB}4.68                                      & \cellcolor[HTML]{CBCEFB}5.44                               & \cellcolor[HTML]{CBCEFB}2.29           & \cellcolor[HTML]{CBCEFB}11.88                             & \cellcolor[HTML]{CBCEFB}1.03           & \cellcolor[HTML]{CBCEFB}1.74           & \cellcolor[HTML]{CBCEFB}1.35           & \cellcolor[HTML]{CBCEFB}0.23           & \cellcolor[HTML]{CBCEFB}2.52           \\ \hline \hline
                                                                                & $A_{\text{er}} $                    & 99.60                                                             & 100.00                                                     & 99.76                                  & 99.88                                                     & 100.00                                 & 100.00                                 & 100.00                                 & 100.00                                 & 99.60                                  \\
                                                                                & $\hat{A}_{\text{adj}}$                   & 92.60                                                             & 95.52                                                      & 94.76                                  & 92.44                                                     & 90.80                                  & 91.28                                  & 97.28                                  & 89.76                                  & 95.40                                  \\ 
\multirow{-3}{*}{\textit{FADE (ours)}}                                          & \cellcolor[HTML]{CBCEFB}ERB        & \cellcolor[HTML]{CBCEFB}\textbf{95.97}                            & \cellcolor[HTML]{CBCEFB}\textbf{97.70}                     & \cellcolor[HTML]{CBCEFB}\textbf{97.19} & \cellcolor[HTML]{CBCEFB}\textbf{96.01}                    & \cellcolor[HTML]{CBCEFB}\textbf{95.17} & \cellcolor[HTML]{CBCEFB}\textbf{95.44} & \cellcolor[HTML]{CBCEFB}\textbf{98.62} & \cellcolor[HTML]{CBCEFB}\textbf{94.60} & \cellcolor[HTML]{CBCEFB}\textbf{97.54} \\ \hline
\end{tabular}
}
% \vspace{-2pt}
\caption{\label{tab:FG_stanford_oxford_cub} \textbf{Evaluation of erasing breeds of dogs, flowers, and birds from the Stanford Dogs, Oxford Flowers, and CUB datasets, respectively.} $A_{\text{er}}$ represents erasing accuracy (higher is better), $\hat{A}_{\text{adj}}$ is the mean adjacency set accuracy (higher is better) from concept neighborhood, and ERB reflects the balance between forgetting and retention.}
\end{table*}

% \begin{equation}
% \text{ERB Score} = \frac{2 \cdot A_{\text{er}} \cdot \left( \frac{1}{|C|} \sum_{c \in C} A_{\text{adj}} \right)}{A_{\text{er}} + \left( \frac{1}{|C|} \sum_{c \in C} A_{\text{adj}} \right)},
% \end{equation}

% \vspace{-2pt}
\section{Experimental Details and Analysis}
% \vspace{-3pt}
\noindent\textbf{Datasets:} We evaluate FADE using two protocols: (a) \textbf{Fine-Grained Unlearning (FG-Un)}, which focuses on erasing $c_{\text{tar}}$ while preserving generalization on challenging concepts in $\mathcal{D}_a$, and (b) \textbf{Coarse-Grained Unlearning (CG-Un)}, which assesses the model’s ability to retain generalization on concepts in $\mathcal{D}_r$. For FG-Un, we utilize fine-grained classification datasets, including Stanford Dogs \cite{stanford_dogs_dataset}, Oxford Flowers \cite{oxford_flowers_dataset}, Caltech UCSD Birds (CUB) \cite{CUB_birds_dataset}, and ImageNet-1k \cite{imagenet_1k_dataset}, due to their closely related classes. We evaluate FADE on three target classes per fine-grained dataset and four target classes in ImageNet-1k. Adjacency sets for these classes are constructed using the Concept Neighborhood. For CG-Un, we follow standard evaluation protocols \cite{esd,receler} for the Imagenette \cite{howard2020fastai} and I2P \cite{schramowski2023safe} datasets, where evaluations focus on the target class and other classes, regardless of semantic similarity.

\noindent\textbf{Baselines:} We compare FADE with state-of-the-art methods for concept erasure, including Erased Stable Diffusion (ESD) \cite{esd}, Concept Ablation (CA) \cite{CA}, Forget-Me-Not (FMN) \cite{FMN}, Semi-Permeable Membrane (SPM) \cite{SPM}, and Receler \cite{receler}. Open-source implementations and standard settings are used for all baseline evaluations.

\noindent\textbf{Evaluation Metrics:} For FG-Un, we measure \textit{Erasing Accuracy} ($A_{\text{er}}$), which quantifies the degree of target concept erasure (higher values indicate better erasure), and \textit{Adjacency Accuracy} ($A_{\text{adj}}$), which evaluates retention across $c \in \mathcal{A}(c_{\text{tar}})$. To balance these, we introduce the \textit{Erasing-Retention Balance (ERB) Score}:

% \vspace{-4pt}
\begin{equation}
\text{ERB Score} = \frac{2 \cdot A_{\text{er}} \cdot \hat{A}_{\text{adj}}}{A_{\text{er}} + \hat{A}_{\text{adj}} + \eta },
\end{equation}

where $\hat{A}_{\text{adj}} = \frac{1}{|C|} \sum_{c \in C} A_{\text{adj}}$ is the mean Adjacency Accuracy, and $\eta$ mitigates divide-by-zero errors. The ERB score provides a harmonic mean to evaluate unlearning and retention balance within $\mathcal{A}(c_{\text{tar}})$. For CG-Un, we follow standard protocols for Imagenette and report classification accuracy from a pre-trained ResNet-50 model before and after unlearning. For I2P, we use NudeNet \cite{bedapudi2019nudenet} to count nudity classes and FID \cite{heusel2017gans} to measure visual fidelity between the original and unlearned models.

\subsection{Results of Fine-Grained Unlearning (FG-Un)}
% \vspace{-2pt}
We evaluate FADE’s fine-grained unlearning performance on Stanford Dogs, Oxford Flowers, and CUB datasets, as shown in Table \ref{tab:FG_stanford_oxford_cub}. We select three target classes for each dataset and define their adjacency sets using Concept Neighborhood with  $K=5$. To address distribution shifts, we fine-tune pre-trained classifiers on each dataset with samples generated by the SD v1.4 model. We then compute Erasing Accuracy ($A_{\text{er}}$) for the erased target class and Adjacency Accuracy ($\hat{A}_{\text{adj}}$), the mean classification accuracy across adjacency set classes  $\mathcal{A}_{\text{adj}}$.

\begin{figure*}[t]
\centering
  \includegraphics[width=0.95\textwidth]{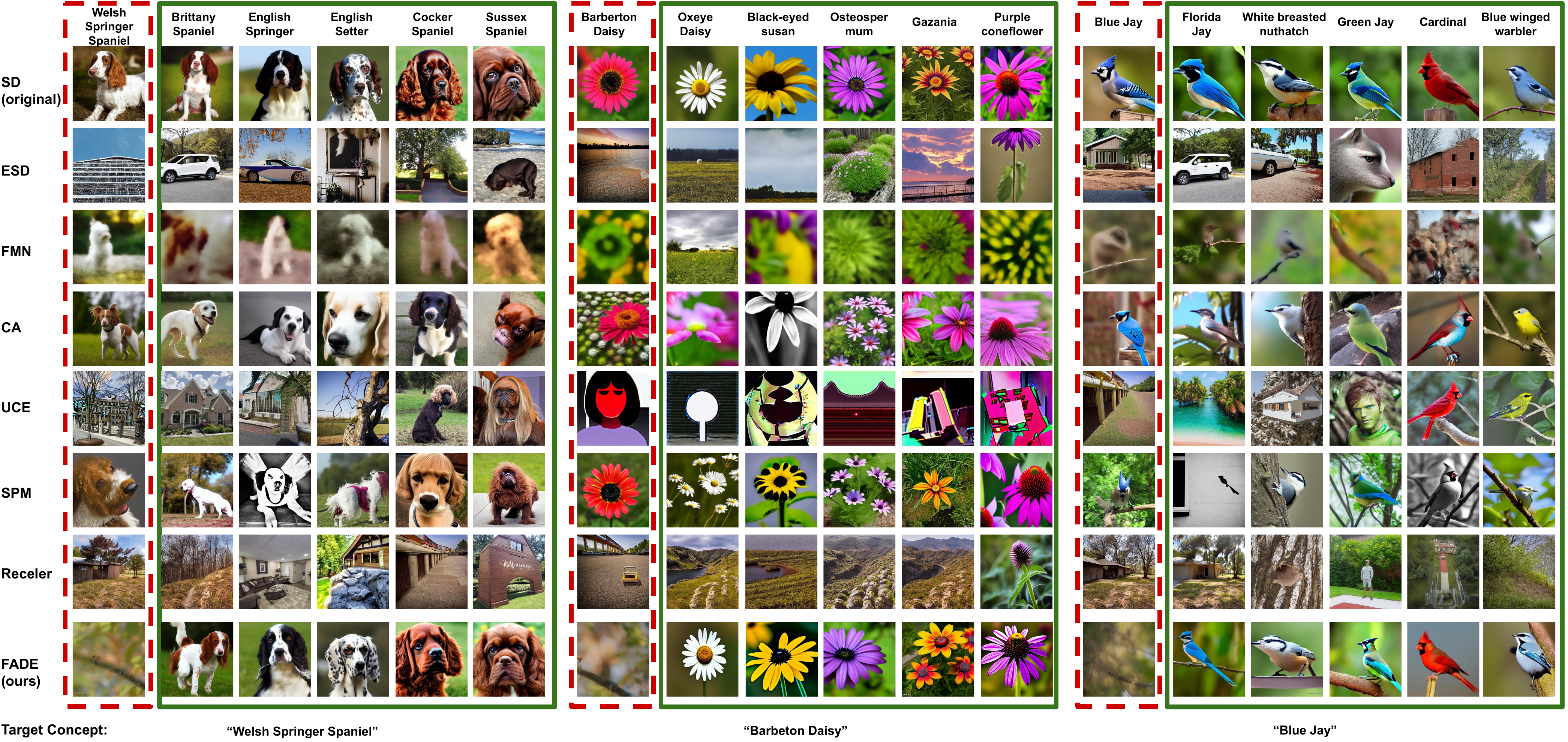}
  % \vspace{-2pt}
\caption{\textbf{Qualitative comparison between existing and proposed algorithms for erasing target concepts and testing retention on neighboring fine-grained concepts}. We visualize one target concept each from the Stanford Dogs, Oxford Flowers, and CUB datasets. Visualizations for more concepts are available in the supplementary.} 
    \label{fig:FG_qualitative}
\end{figure*}

\begin{figure}[ht]
%\vspace{-15pt}
\centering
  \includegraphics[scale=0.6]{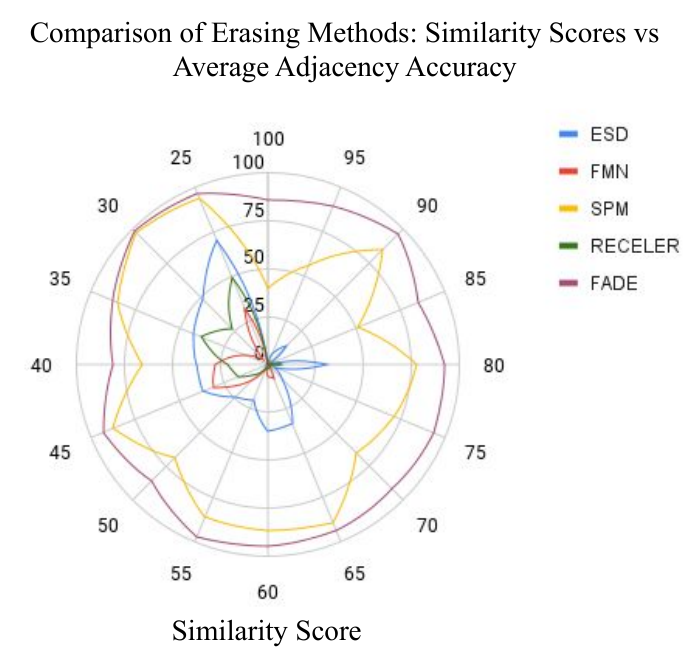}
\caption{Radar plot comparing FADE with existing unlearning methods (ESD, FMN, SPM, Receler) by structural similarity score (circular axis, \%) and adjacency accuracy (radial axes) on concepts from the ImageNet-1k dataset. Most methods begin to degrade beyond a similarity score of 70\%, with SPM resilient until 90\% and FADE showing the highest robustness. For fair analysis, only methods with $A_{\text{er}} \leq 20\%$ are considered.}
    \label{fig:IA}
\end{figure}

% \footnotetext{For a fair analysis, methods with $A_{\text{er}} \leq 20\%$ are considered.}

% Comparison Table on Imagenette dataset for CG-Un.
\begin{table*}[t]
\centering
\resizebox{0.95\linewidth}{!}{%
\begin{tabular}{l|cc|cc|cc|cc|cc|cc|cc}
\hline
\multirow{2}{*}{} & \multicolumn{2}{c|}{Original SD v1.4}      & \multicolumn{2}{c|}{ESD}                   & \multicolumn{2}{c|}{FMN}                   & \multicolumn{2}{c|}{CA}                    & \multicolumn{2}{c|}{SPM}                   & \multicolumn{2}{c|}{Receler}               & \multicolumn{2}{c}{FADE (ours)}            \\ \cline{2-15} 
                  & $A_{\text{tar}} \uparrow$                    & $\hat{A}_{\text{others}} \uparrow$                    & $A_{\text{tar}} \downarrow$ & $\hat{A}_{\text{others}} \uparrow$ & $A_{\text{tar}} \downarrow$ & $\hat{A}_{\text{others}} \uparrow$ & $A_{\text{tar}} \downarrow$ & $\hat{A}_{\text{others}} \uparrow$ & $A_{\text{tar}} \downarrow$ & $\hat{A}_{\text{others}} \uparrow$ & $A_{\text{tar}} \downarrow$ & $\hat{A}_{\text{others}} \uparrow$ & $A_{\text{tar}} \downarrow$                 & $\hat{A}_{\text{others}} \uparrow$                \\ \hline
Cassette Player   & 25.00                               & 87.58                                      &         0.60         &    65.50                     & 4.00             & 20.93                   & 20.20            & 85.35                   & 2.00             & 87.31                   & 0.00             & 77.08                   & 0.00                             & 86.28                                  \\
Chain Saw         & 64.00                               & 90.52                                      & 0.00             & 66.66                   & 0.00             & 39.22                   & 72.80            & 86.35                   & 20.22            & 81.44                   & 0.00             & 70.22                   & 0.00                             & 88.90                                  \\
Church            & 82.00                               & 88.27                                      &       0.10           &    69.88                     & 4.00             & 52.73                   & 47.00            & 83.64                   & 78.0             & 87.15                   & 0.80             & 72.93                   & 0.00                             & 85.15                                  \\
French Horn       & 99.8                                & 88.55                                      &          0.20        &   60.55                        & 3.00             & 38.13                   & 100.00           & 86.11                   & 13.89            & 76.91                   & 0.00             & 66.37                   & 0.00                             & 87.22                                  \\
Gas Pump          & 81.85                               & 89.7                                       &         4.0         &              62.71           & 0.87             & 39.97                   & 90.60            & 86.67                   & 16.00            & 80.26                   & 0.00             & 66.57                   & 0.00                             & 89.30                                  \\
Parachute         & 97.24                               & 86.37                                      &         4.0         &  72.67                         & 11.60            & 54.33                   & 94.39            & 85.88                   & 53.60            & 82.55                   & 1.00             & 72.57                   & 0.72                             & 84.05                                  \\
Tench             & 72.00                               & 88.23                                      & 0.00             & 72.22                   & 1.79             & 56.22                   & 68.80            & 86.26                   & 21.80            & 81.46                   & 3.00             & 7.66                    & 0.00                             & 87.85                                  \\
English Springer  & 97.00                               & 86.40                                      &       5.2           &          68.57               & 9.40             & 65.57                   & 35.50            & 79.11                   & 38.88            & 81.11                   & 47.88            & 76.26                   & 0.00                             & 83.75                                  \\
Garbage Truck     & 94.64                               & 89.525                                     &       0.80           &         62.57                & 0.27             & 21.26                   & 75.00            & 83.86                   & 27.20            & 79.42                   & 0.00             & 64.97                   & 0.00                             & 88.20                                  \\
Golf Ball         & 99.85                               & 86.05                                      &      0.60            &             61.50            & 24.00            & 54.80                   & 99.60            & 86.08                   & 50.00            & 81.75                   & 0.00             & 69.82                   & 1.6                              & 85.25                                  \\ \hline
\end{tabular}
}
\caption{\label{tab:CG_imagenet}\textbf{Comparative evaluation for coarse-grained unlearning on the Imagenette dataset with existing state-of-the-art methods.} For all models except the original SD model, a lower $A_{\text{tar}}$ indicates better erasure of the target concept, and a higher $\hat{A}_{\text{others}}$ represents better retention, as it is the average accuracy on the non-targeted concepts. Except for ‘Cassette Player,’ $\hat{A}_{\text{others}}$ is computed over 8 classes, excluding it due to its lower original accuracy for consistency with prior work.}
\end{table*}

% Comparospn table on FG-Un on ImageNet-1k dataset.
\begin{table}[t]
\centering
% \vspace{-3pt}
\resizebox{\linewidth}{!}{%
\begin{tabular}{l|cccc}
\hline
Target Class & \multicolumn{1}{l}{Golf Ball} & \multicolumn{1}{l}{Garbage Truck} & \multicolumn{1}{l}{English Springer} & \multicolumn{1}{l}{Tench} \\ \hline
ESD          & 44.81                         & 44.91                             & 50.07                                & 74.74                     \\
FMN          & 49.62                         & 3.30                               & 1.42                                 & 56.96                     \\
CA           & 0.79                          & 39.49                             & 63.56                                & 35.72                     \\
SPM          & 63.64                         & 75.18                             & 86.02                                & 72.30                      \\
Receler      & 20.07                         & 32.77                             & 47.62                                & 56.36                     \\ \hline 
FADE (ours)  & \textbf{96.82}                         & \textbf{91.65}                             & \textbf{97.93}                                & \textbf{87.08}                     \\ \hline
\end{tabular}
}
% \vspace{-2pt}
\caption{\label{tab:FG_imagenet}\textbf{Evaluation of erasing structurally similar concepts from ImageNet-1k dataset.} We present the ERB scores, with FADE significantly outperforming all existing algorithms. $A_{\text{er}}$ and $\hat{A}_{\text{adj}}$ are available in the supplementary.}
\end{table}

\begin{figure*}[t]
\centering
  \includegraphics[width=0.95\textwidth]{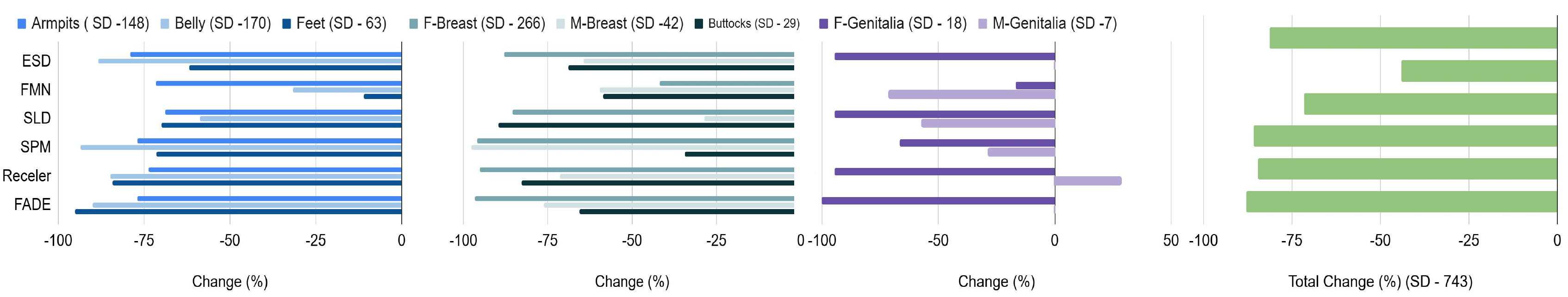}
\caption{\textbf{NudeNet Evaluation on the I2P benchmark.} The numbers followed by "SD" indicate the count of exposed body parts in the SD v1.4 generations. The binplots show the reduction achieved by different methods for erasing nudity. Compared to prior works, FADE effectively eliminates explicit content across various nude categories.} 
    \label{fig:I2P_results}
\end{figure*}

\begin{figure*}[t]
\centering
  \includegraphics[width=\linewidth]{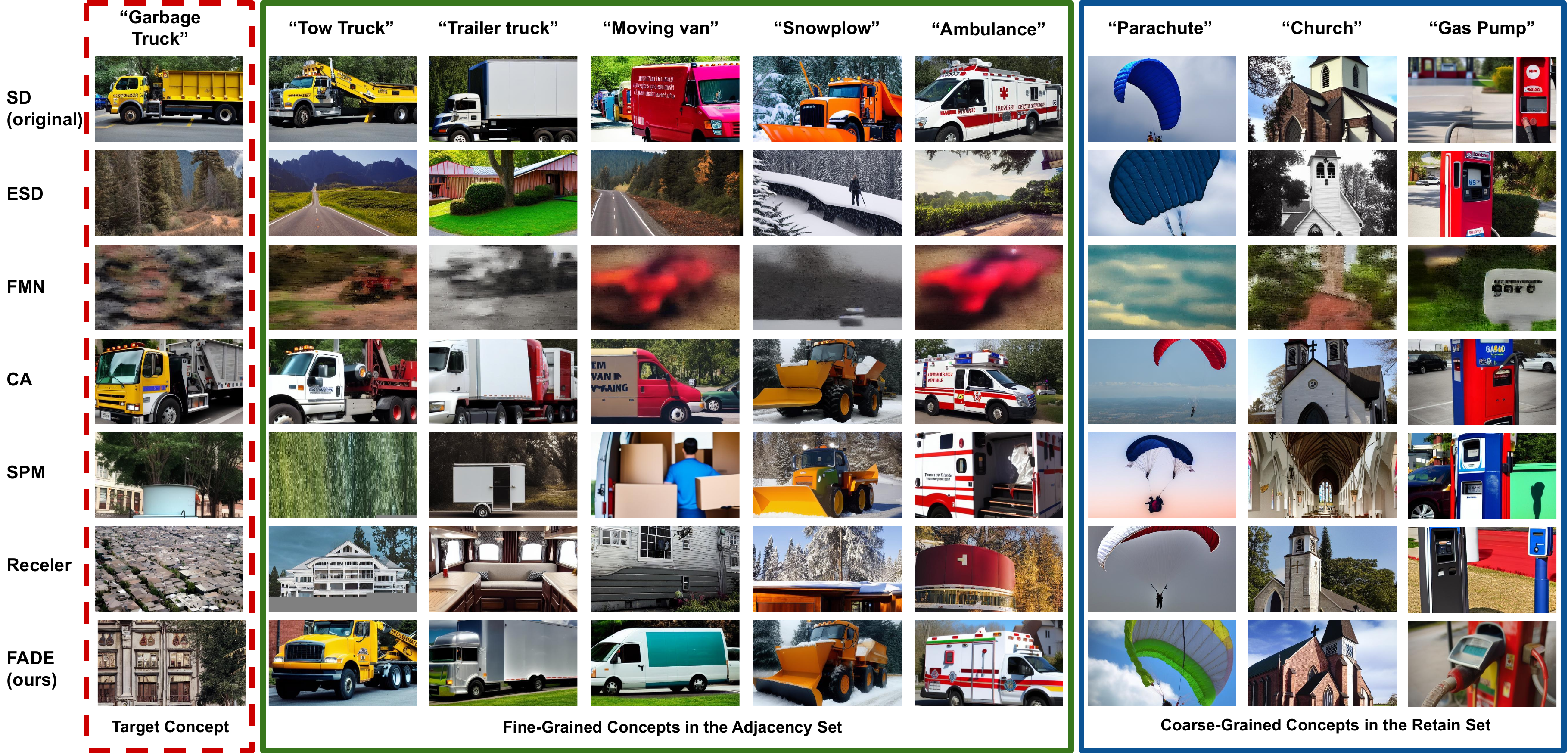}
\caption{Comparison of FADE with various algorithms for erasing the 'garbage truck' class in Fine-Grained and Coarse-Grained Unlearning. The target class, adjacency set and the retain set and constructed from the ImageNet-1k dataset.} 
    \label{fig:CG_qualitative}
\end{figure*}

\begin{figure}[ht]
\centering
%\vspace{-5pt}
  \includegraphics[width=\linewidth]{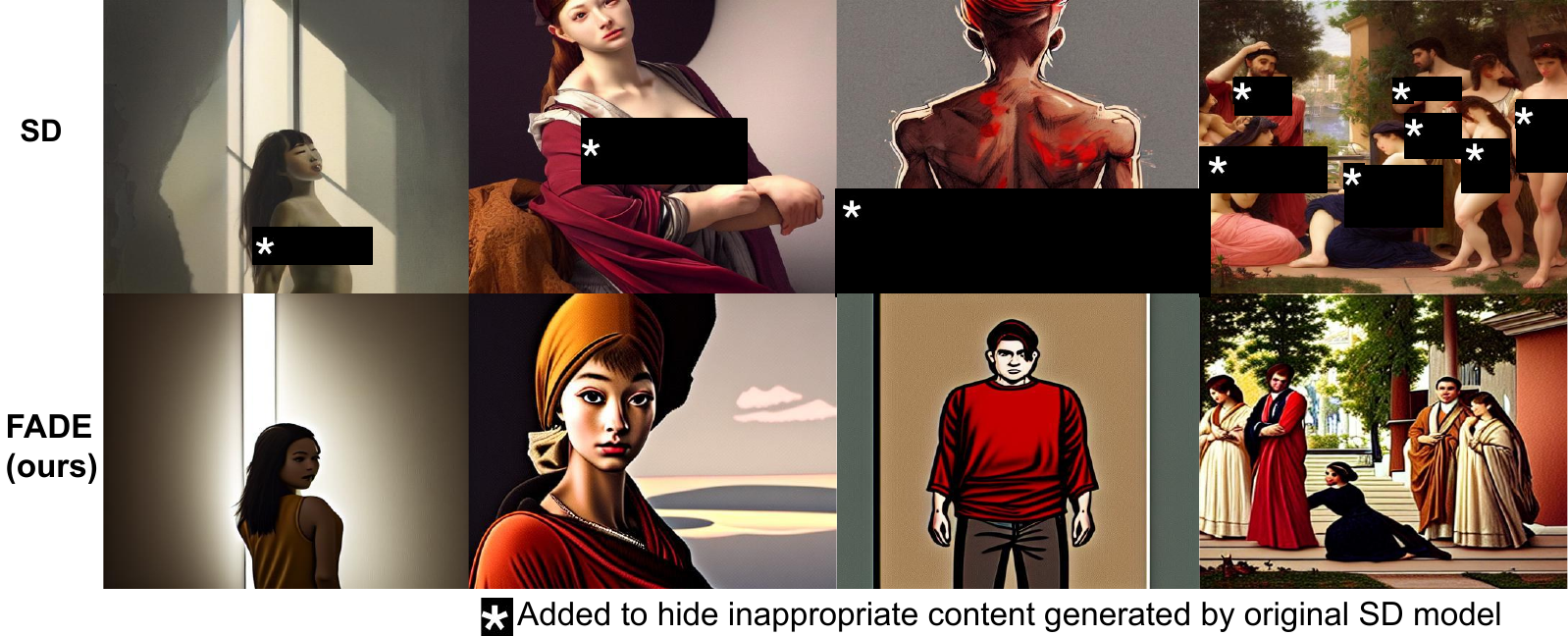}
\caption{Visualization of before and after unlearning nudity through FADE. The prompts are borrowed from I2P dataset. } 
    \label{fig:I2P_qualitative}
\end{figure}

\noindent \textbf{Performance on Fine-Grained Datasets:}  
Table \ref{tab:FG_stanford_oxford_cub} demonstrates that existing algorithms struggle to retain neighboring concepts while erasing the target concept, as reflected by their low ERB scores. FMN shows the weakest adjacency retention, followed by Receler and ESD. CA and SPM perform moderately, but FADE consistently outperforms all baselines by at least 15\% across all target classes. This highlights FADE’s superior ability to balance effective erasure of $c_{\text{tar}}$ with the retention of adjacent classes in $\mathcal{A}(c_{\text{tar}})$, showcasing its effectiveness in fine-grained unlearning tasks. Further evaluations for top-10 adjacency concepts (available in supplementary) shows the effectiveness of FADE for FG-Un.

% \vspace{-2pt}
\noindent \textbf{Qualitative Analysis:}  
Figure \ref{fig:FG_qualitative} presents generation results for one target class and its adjacency set from each dataset before and after applying unlearning algorithms (additional examples in supplementary material). The first row shows images generated by the original SD model, followed by results from each unlearning method. Consistent with Table \ref{tab:FG_stanford_oxford_cub}, ESD, FMN, and Receler fail to preserve fine-grained details of neighboring classes. CA and SPM retain general structural features but often struggle with specific attributes like color in dog breeds (e.g., Brittany Spaniel, Cocker Spaniel), bird species (e.g., Florida Jay, Cardinal), and flower species. These methods frequently produce incomplete erasure or generalized representations. In contrast, FADE preserves fine-grained details while ensuring effective erasure, as evidenced by sharper distinctions in adjacency sets.

\noindent \textbf{Evaluation on ImageNet-1k Dataset:}  
FADE’s performance on ImageNet-1k is evaluated for target classes such as Balls, Trucks, Dogs, and Fish. Using Concept Neighborhood, adjacency sets closely align with the manually curated fine-grained class structure by Peychev et al. \cite{peychev2023automated}, demonstrating Concept Neighborhood’s accuracy. Table \ref{tab:CG_imagenet} shows that FADE outperforms all baselines, achieving at least 12\% higher ERB scores than SPM, the next-best method. FMN and CA perform poorly in both adjacency retention and erasure. Additional details on adjacency composition and metrics are provided in the supplementary.

\noindent \textbf{Adjacency Inflection Analysis:}  
We evaluate the robustness of algorithms as semantic similarity increases using fine-grained classes from ImageNet-1k and fine-grained datasets. Figure \ref{fig:IA} illustrates the relationship between CLIP-based structural similarity (circular axis, \%) and average adjacency accuracy (radial axes). FMN and ESD degrade at 78\% similarity, with Receler failing at 80\%. SPM shows moderate resilience but struggles beyond 90\% similarity. In contrast, FADE maintains high adjacency accuracy, demonstrating robustness even at high similarity levels, validating its effectiveness in adjacency-aware unlearning tasks.

\subsection{Coarse-Grained Unlearning(CG-Un) Results}
% \vspace{-2pt}
We evaluate FADE and state-of-the-art methods on the Imagenette dataset, which exhibits minimal semantic overlap. Results are presented in Table \ref{tab:CG_imagenet}. For each target class, we report the target erasure accuracy ($A_{\text{tar}}$, lower is better) and the average accuracy on other classes ($\hat{A}_{others}$, higher is better). These metrics assess erasure on $\mathcal{D}_u$ and retention on $\mathcal{D}_r$. FADE achieves the best balance between erasure and retention, outperforming all baselines. CA and SPM perform moderately well due to their partial target class removal, which preserves structure and enhances retention. Receler, ESD, and FMN exhibit sub-optimal performance, with FMN being the weakest.

\noindent \textbf{Qualitative Analysis:}  
Figure \ref{fig:FG_qualitative} illustrates qualitative results for the overlapping class of “Garbage Truck” from ImageNet-1k and Imagenette. While ESD, FMN, SPM, and Receler unlearn the target class, they struggle with generalizability across adjacent classes. FADE, in contrast, demonstrates robust generalizability in both FG-Un (ImageNet-1k) and CG-Un (Imagenette), achieving the highest overall performance (Table \ref{tab:CG_imagenet}). Additional visualizations for FG-Un and CG-Un classes are included in the supplementary.

\noindent \textbf{Nudity Erasure on I2P:}  
We further evaluate FADE on I2P nudity prompts using NudeNet to detect targeted nudity classes. FADE achieves the highest erasure ratio change of 87.88\% compared to the baseline SD v1.4 model, outperforming all methods. Among competitors, SPM ranks second, followed by Receler and ESD. On the nudity-free COCO30K dataset, FADE scores an FID of 13.86, slightly behind FMN (13.52). However, FMN’s erasure ratio change is significantly lower at 44.2\%, highlighting its ineffectiveness in nudity erasure. Figure \ref{fig:I2P_qualitative} shows qualitative results, illustrating FADE’s superior performance in removing nudity across various prompts.

\begin{table}[!t]
% \footnotesize
\centering
\begin{tabular}{lll|lll}
\hline
% \lambda_{\text{exp}} \mathcal{L}_{\text{exp}} + \lambda_{\text{adj}} \mathcal{L}_{\text{adj}} + \lambda_{\text{guid}} \mathcal{L}_{\text{guid}}
\multicolumn{3}{c|}{Components} & \multicolumn{3}{c}{Metrics} \\
                         $\mathcal{L}_{\text{guid}}$ &  $\mathcal{L}_{\text{er}}$ & $\mathcal{L}_{\text{adj}}$ & $A_{\text{er}}$   $\uparrow$      &    $\hat{A}_{\text{adj}}$  $\uparrow$   &      ERB $\uparrow$  \\ \hline
$\checkmark$ & $\checkmark$ & $\checkmark$ & 99.60        & 92.60         &    95.97     \\
$\checkmark$ & $\checkmark$ & $\times$ & 25.40        & 80.24         &    38.58    \\
$\checkmark$ & $\times$ & $\checkmark$ & 28.00      & 95.08         &    43.26   \\
$\checkmark$ & $\times$ & $\times$ & 31.80        & 78.16        &    45.20    \\
$\times$ & $\checkmark$ & $\checkmark$ & 100.0        & 76.12       &    86.44   \\
$\times$ & $\checkmark$ & $\times$ & 43.60       & 90.44        &    58.83   \\
 \hline
 
\end{tabular}
\caption{\label{tab:ablation}Ablation study with different components of FADE with target class as Welsh Springer Spaniel.}
\end{table}

\subsection{Ablation Study}
% \vspace{-2pt}
We study the individual contributions of FADE’s loss components: guidance loss ($\mathcal{L}_{\text{guid}}$), erasing loss ($\mathcal{L}_{\text{er}}$), and adjacency loss ($\mathcal{L}_{\text{adj}}$). Table \ref{tab:ablation} shows results for the target class “Welsh Springer Spaniel” using Erasure Accuracy ($A_{\text{er}}$), Adjacency Accuracy ($\hat{A}_{\text{adj}}$), and the ERB score. The complete model achieves the highest ERB score of 95.97, balancing target erasure and adjacency preservation. Excluding $\mathcal{L}_{\text{adj}}$ results in a sharp drop to 38.58 ERB, highlighting its role in adjacency retention. Removing $\mathcal{L}_{\text{er}}$ reduces ERB to 43.26, emphasizing its importance in precise erasure. Similarly, omitting $\mathcal{L}_{\text{guid}}$ achieves perfect erasure accuracy (100.0) but lowers ERB to 86.44, reflecting its necessity for maintaining structural integrity in the adjacency~set. %These findings confirm the critical interplay of all three components in enabling FADE to achieve robust, adjacency-aware unlearning.

%\section{Discussion, Limitations, and Conclusion}
%In this paper, we investigate a phenomenon introduced by unlearning in text-to-image generative foundation models, which we term as \textit{adjacency}. Our findings reveal that concepts proximate to the target concept in the learned manifold are disproportionately impacted by unlearning processes. This is largely due to the approximate nature of current unlearning algorithms, which often rely on feature displacement to achieve erasure. This displacement, while effective in eliminating target concepts, deforms the manifold in ways that compromise the integrity of semantically similar concepts. True unlearning, in theory, would require the selective removal of knowledge, akin to creating “holes” in the semantic space, yet achieving this in a continuous manifold remains an open challenge. We leave this as a compelling direction for future research.

%Despite FADE’s effectiveness in fine-grained unlearning, it has limitations. Specifically, erased concepts can still emerge through alternative or adversarial prompts, demonstrating the need for more robust, context-aware unlearning methods to address such circumvention.

%In conclusion, FADE introduces adjacency-aware erasure via Concept Lattice and a Mesh-based fading approach. Extensive experiments demonstrate its ability to effectively erase target concepts while preserving adjacent and unrelated knowledge, advancing generative unlearning and emphasizing the critical role of adjacency in maintaining model fidelity.

\subsection{Qualitative and User Study}  
% \vspace{-2pt}
We conducted a user study with 40 participants aged 18–89 to evaluate FADE's performance from a human perspective. Participants assessed both erasure and retention tasks across nine target concepts (see Table \ref{tab:FG_stanford_oxford_cub}), each paired with their top three related concepts from the Stanford Dogs, Oxford Flowers, and CUB datasets. For the \textit{erasure evaluation}, participants judged whether images generated by the unlearned models effectively removed the target concept. For the \textit{retention evaluation}, they assessed if adjacent classes were correctly retained. Real examples were provided beforehand to ensure consistency. Each participant evaluated 81 images. Scores from the erasure and retention tasks were aggregated to compute the ERB score for each method. The user study results yielded ERB scores as follows: FADE achieved the highest score of 59.49, outperforming CA (49.38), SPM (49.13), FMN (43.07), ESD (38.43), and Receler (0.06). Participants noted that CA often failed to fully remove the target concept, while Receler adversely affected adjacent classes. These findings highlight FADE's superiority in balancing effective erasure and retention, as perceived by human evaluators.

\section{Conclusion}
% \vspace{-4pt}
This work introduces \textit{adjacency} in unlearning for text-to-image models, highlighting how semantically similar concepts are disproportionately affected during erasure. Current algorithms rely on feature displacement, which effectively removes target concepts but distorts the semantic manifold, impacting adjacent concepts. Achieving fine-grained unlearning, akin to creating ``holes'' in the manifold, remains an open challenge. The proposed FADE effectively erases target concepts while preserving adjacent knowledge through the Concept Neighborhood and Mesh modules. FADE advances adjacency-aware unlearning, emphasizing its importance in maintaining model fidelity.

% \vspace{-2pt}
\section{Acknowledgement}
% \vspace{-4pt}
The authors thank all volunteers in the user study. This research is supported by the IndiaAI mission and Thakral received partial funding through the PMRF~Fellowship.

%-------------------------------------------------------------------------

% \input{sec/X_suppl}

% % \input{sec/2_formatting}
% % \input{sec/3_finalcopy}

% \input{sec_rev/0_abstract}    
% \input{sec_rev/1_intro}
% \input{sec_rev/2_formatting}
% \input{sec_rev/3_finalcopy}

{
    \small
    \bibliographystyle{ieeenat_fullname}
    \bibliography{main.bib}
}

\clearpage
\setcounter{page}{1}
\maketitlesupplementary

\section{Theoretical Basis for Concept Lattice}
\label{sec:rationale}
Based on the literature, as noted in the work from Duda et al. \cite{duda2001pattern}, we extend the observations of Boiman et al. \cite{boiman2008defense} as a theoretical justification for the proposed nearest neighbor-based concept lattice, which approximates the gold-standard Naive Bayes classifier for constructing the adjacency set. \\

\iffalse
\noindent\textbf{Theorem: k-Nearest Neighbor in $\mathbb{R}^d$ as an approximation to Naive Bayes.}

\noindent Let  $\mathbf{x} \in \mathbb{R}^{h \times w \times c}$  be an image, with  height, width, and number of channels as $h$,  $w$, and $c$, respectively. Let  $\phi: \mathbb{R}^{h \times w \times c} \to \mathbb{R}^d$  be the mapping function of $\mathbf{x}$ to a latent feature space $\mathbb{R}^d$, where  $d \ll hwc$. Assume that the latent features  $z := \phi(\mathbf{x})$  are conditionally independent given the class label  $C \in \mathcal{C}$. As the sample size  $N \to \infty$  and the number of neighbors  $k \to \infty$  with  $k/N \to 0$, the k-Nearest Neighbors (k-NN) classifier operating in $\mathbb{R}^d$ converges to the Bayes optimal classifier, which, under the conditional independence assumption, is the Naive Bayes classifier. Formally,

\begin{equation*}
\lim_{N \to \infty} P(C_{\text{K-NN}}(\phi(\mathbf{x})) = C_{\text{NB}}(\mathbf{x})) = 1.    
\end{equation*}
\fi

\vspace{-14pt}
\noindent\rule{\linewidth}{0.4pt}
\vspace{-10pt}

% \begin{theorem}[k-NN Approximation to Naive Bayes in $\mathbb{R}^d$]
\noindent\textbf{Theorem 1} (k-NN Approximation to Naive Bayes in $\mathbb{R}^d$)
Let $\mathbf{x} \in \mathbb{R}^{h \times w \times c}$ represent an image with dimensions height $h$, width $w$, and channels $c$. Let the mapping function 
$\phi: \mathbb{R}^{h \times w \times c} \to \mathbb{R}^d$ project the image $\mathbf{x}$ into a latent feature space $\mathbb{R}^d$, where $d \ll hwc$. Assume that the latent features $z := \phi(\mathbf{x})$ are conditionally independent given the class label $C \in \mathcal{C}$. 

Then, the k-Nearest Neighbors (k-NN) classifier operating in $\mathbb{R}^d$ converges to the Naive Bayes classifier as the sample size $N \to \infty$, the number of neighbors $k \to \infty$, and $k/N \to 0$. Specifically,
\begin{equation}
\lim_{N \to \infty} P\bigl(C_{\text{k-NN}}(\phi(\mathbf{x})) = C_{\text{NB}}(\mathbf{x})\bigr) = 1.
\end{equation}
% \end{theorem}

% \\ \noindent\textbf{Preliminaries and Notations}

\noindent\textbf{Proof Outline:}
\noindent Let  $D = \{(\mathbf{x}_i, y_i)\}_{i=1}^N$  be a dataset consisting of images  $\mathbf{x}_i \in \mathbb{R}^{h \times w \times c}$  and their corresponding class labels  $y_i \in \mathcal{C}$. Each image  $\mathbf{x}_i$  is mapped to a latent space  $\mathbb{R}^d$  through the mapping function  $\phi: \mathbb{R}^{h \times w \times c} \to \mathbb{R}^d$, resulting in a latent feature vector  $z_i := \phi(\mathbf{x}_i)$.

\noindent We assume the following:
\begin{itemize}
    \item The latent feature vectors  $\phi(\mathbf{x})$  are conditionally independent given the class label  $y$.
    \item The representation function $\phi(\mathbf{x})$  preserves the class-conditional structure in $\mathbb{R}^d$, such that images of the same class remain clustered in proximity to one another.
    \item $d$  is sufficiently large ensuring high separability between classes while remaining lower-dimensional than the original input space, i.e., in $d \ll hwc$.
\end{itemize}

\noindent\textbf{Proof:} We follow the outline above proceeding step by step.

\noindent\textbf{Step 1: Bayes Optimal Classifier (Naive Bayes)}
\noindent The Bayes optimal classifier is defined as the classifier that minimizes the expected classification error by choosing the class that maximizes the posterior probability  $P(C = c | \mathbf{x})$. Under the Naive Bayes assumption, the posterior decomposes as follows:

\begin{equation}
    P(C = c | \mathbf{x}) = \frac{P(\mathbf{x} | C = c) P(C = c)}{P(\mathbf{x})}.    
\end{equation}

Given the conditional independence of $\mathbf{z}$ in $\mathbb{R}^d$, the class-conditional likelihood  $P(\mathbf{x} | C = c)$  is factorized over the components of the latent vector  $\phi(\mathbf{x}) = (\phi_1(\mathbf{x}), \ldots, \phi_d(\mathbf{x}))$ , i.e.,

\begin{equation}
    P(\mathbf{x} | C = c) = \prod_{j=1}^d P(\phi_j(\mathbf{x}) | C = c).
\end{equation}

Thus, the decision rule of the Naive Bayes classifier becomes:

\begin{equation}
    C_{\text{NB}}(\mathbf{x}) = \arg\max_{c \in \mathcal{C}} P(C = c) \prod_{j=1}^d P(\phi_j(\mathbf{x}) | C = c).
\end{equation}

\noindent\textbf{Step 2: k-Nearest Neighbor Classifier in $\mathbb{R}^d$}

\noindent The k-NN classifier operates in the latent space $\mathbb{R}^d$, assigns a class label $C_{\text{k-NN}}$ to a query vector $\phi(\mathbf{x})$ by selecting the nearest instance in $\mathbb{R}^d$. For two images $\mathbf{x}$  and  $\mathbf{x}_i$, we can defined it formally as:

\begin{equation}
    C_{\text{k-NN}} = \arg\max_{i} \text{sim}(\phi(\mathbf{x}), \phi(\mathbf{x}_i)) = \frac{\langle \phi(\mathbf{x}), \phi(\mathbf{x}_i) \rangle}{\|\phi(\mathbf{x})\| \|\phi(\mathbf{x}_i)\|}.
\end{equation}

The k-NN classifier assigns the label to the query image $\mathbf{x}$  by aggregating the labels of its k -nearest neighbors  $\mathcal{N}_k(\mathbf{x})$ in the latent space. This is formally described as:

\begin{equation}
    C_{\text{k-NN}}(\phi(\mathbf{x})) = \arg\max_{c \in \mathcal{C}} \sum_{\mathbf{x}_i \in \mathcal{N}_k(\mathbf{x})} \mathbb{I}(y_i = c),
\end{equation}

where  $\mathbb{I}(y_i = c)$  is the indicator function, returning 1 if  $y_i = c$  and $0$ otherwise. \\

\noindent\textbf{Step 3: Convergence of k-NN to Bayes Optimal~Classifier}\\
\noindent As established by Covert \& Hart et al. \cite{cover1967nearest} in statistical learning theory, the k-NN classifier converges to the Bayes optimal classifier as $N \to \infty$, provided that $k \to \infty$  and  $k/N \to 0$. That is, for sufficiently large $N$ and $k$, the decision rule of the k-NN classifier approximates that of the Bayes optimal classifier $C_{\text{Bayes}(\mathbf{x})}$, i.e.,

\begin{equation}
\lim_{N \to \infty} P(C_{\text{K-NN}}(\phi(\mathbf{x})) = C_{\text{Bayes}}(\mathbf{x})) = 1.
\end{equation}

This convergence holds because, with increase in $N$, $\mathcal{N}_k(\mathbf{x})$ increasingly reflects the local distribution of data around $\mathbf{x}$, which aligns with the underlying class-conditional probability distribution. \\

\noindent\textbf{Step 4: Consistency of k-NN with Naive Bayes in $\mathbb{R}^d$}

\noindent Given the Naive Bayes assumption that the components  $\phi_j(\mathbf{x})$  of the latent representation  $\phi(\mathbf{x})$  are conditionally independent given the class label, the Bayes optimal classifier in this latent space is precisely the Naive Bayes classifier  $C_{\text{NB}}(\mathbf{x})$ . Therefore, we have:

\begin{equation}
    C_{\text{Bayes}}(\mathbf{x}) = C_{\text{NB}}(\mathbf{x}),
\end{equation}

where $C_{\text{Bayes}}(\mathbf{.})$ operates on the latent representations  $\phi(\mathbf{x})$. Combining equation 7 with the equation 6, we conclude that:

\begin{equation}
    \lim_{N \to \infty} P(C_{\text{K-NN}}(\phi(\mathbf{x})) = C_{\text{NB}}(\mathbf{x})) = 1.
\end{equation}

This establishes that the CLIP-based K-NN classifier converges to the Naive Bayes classifier as the sample size grows, provided the assumptions of conditional independence hold in the latent space $\mathbb{R}^d$.

\noindent\textbf{Remarks:}
\begin{itemize}
    \item In high-dimensional spaces, Bayers et al. \cite{beyer1999nearest} proposed concentration of distances implying that Euclidean distance and Cosine similarity perform similarly as $d \to \infty$, ensuring that the use of cosine similarity in latent space provides robust distance-based classification.
    \item In our implementation, the mapping function $\phi: \mathbb{R}^{h \times w \times c} \to \mathbb{R}^d$ is a pre-trained CLIP model, serving for dimensionality reduction where $d \ll hwc$.
    \item The CLIP model's latent space captures abstract and semantic features, reducing the dependency between the components of $\phi(\mathbf{x})$. This makes the assumption of conditional independence more plausible in $\mathbb{R}^d$, allowing Naive Bayes to model the class-conditional likelihoods accurately in the latent space.
    \item For k-NN to converge to optimal Bayes classifier, k must satisfy $k \to \infty$ and $N \to \infty$.    
\end{itemize}
\vspace{-4pt}
\noindent\rule{\linewidth}{0.4pt}
\vspace{-15pt}

% \begin{figure}[]
% %\vspace{-15pt}
% \centering
%   \includegraphics[scale=0.62]{Images/IA.pdf}
% \caption{\textbf{Radar plot comparing FADE with existing unlearning methods (ESD, FMN, SPM, Receler).} For a fair analysis, methods with $A_{\text{er}} \leq 20\%$ are considered. The plot shows structural similarity scores (circular axis, \%) and adjacency accuracy (radial axes) on concepts from the ImageNet-1k dataset. Most methods begin to degrade beyond a similarity score of 70\%, with SPM remaining resilient until 90\% and FADE demonstrating the highest robustness.} 
%     \label{fig:IA}
% \end{figure}

\section{Adjacency Inflection Analysis}
This section examines the breaking point of existing algorithms in preserving adjacency—specifically, at what similarity threshold these methods begin to fail. To evaluate robustness, we analyze the performance of each algorithm as semantic similarity increases, using fine-grained classes from ImageNet-1k and other fine-grained datasets. Figure \ref{fig:IA} illustrates the relationship between CLIP-based semantic similarity (circular axis, \%) and average adjacency accuracy (radial axes).

Results show that FMN and ESD degrade significantly at 78\% similarity, while Receler fails at 80\%. Although SPM demonstrates moderate resilience, it begins to falter beyond 90\% similarity, marking a critical threshold where all existing methods fail to preserve adjacency effectively. In stark contrast, FADE maintains high adjacency accuracy even at elevated similarity levels, demonstrating superior robustness. These findings validate FADE’s efficacy in adjacency-aware unlearning, outperforming state-of-the-art approaches under challenging fine-grained conditions.

% \footnotetext{For a fair analysis, methods with $A_{\text{er}} \leq 20\%$ are considered.}

\section{Adjaceny Retention Analysis}
During training, FADE explicitly considers the top-$k$ adjacent classes (with $k=5$ in all experiments). However, to ensure FADE's generalization beyond the explicitly trained adjacent classes, we evaluate its performance on unseen adjacent concepts (i.e., classes with rank $>5$).

\begin{figure}[t]
\centering
  \includegraphics[width=\linewidth]{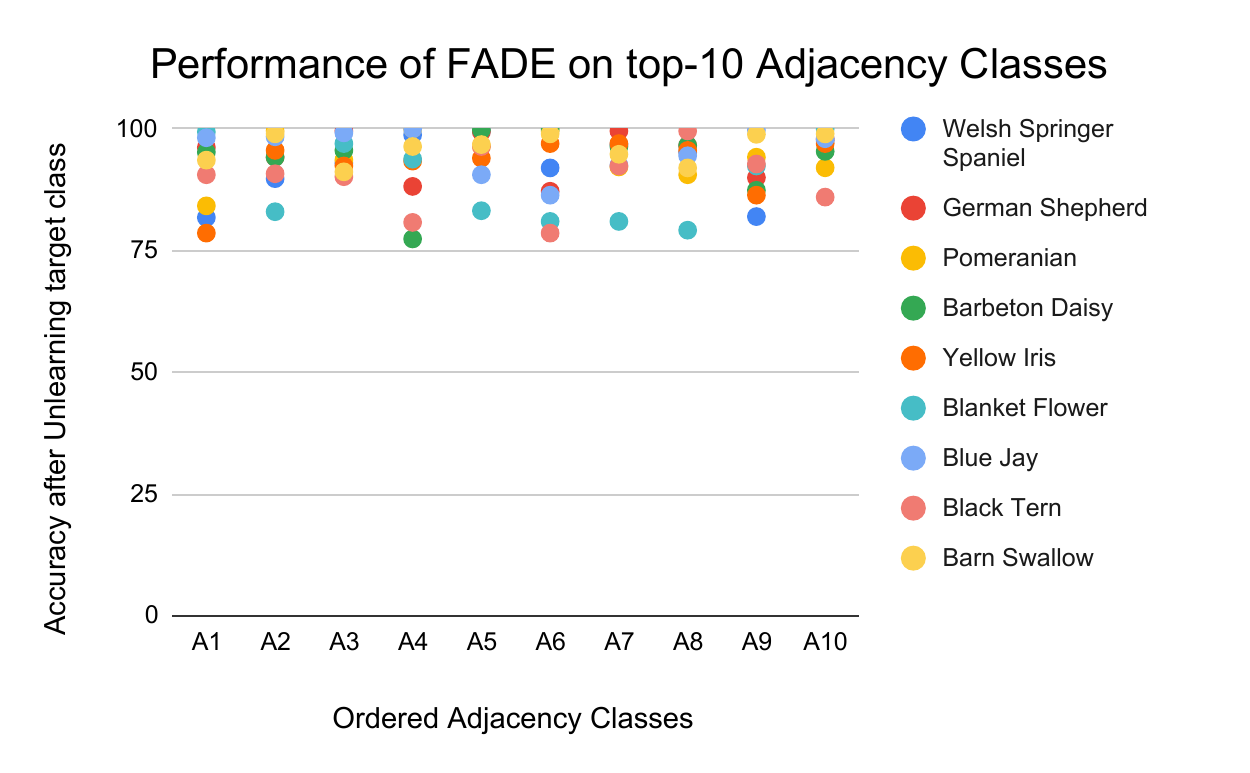}
\caption{For each target class in Table \ref{tab:similarity_scores}, we illustrate the performance of FADE on the top-10 adjacent classes. Adjacent classes are ordered by similarity scores. It is observed that FADE generalizes well on all adjacent classes after unlearning the target class.} 
    \label{fig:top-10}
\end{figure}

We assess FADE’s adjacency retention by analyzing classification accuracy across the top-10 adjacent classes for each target concept (as detailed in Table \ref{tab:similarity_scores}). Using classifiers trained on their respective datasets, we measure retention accuracy for Stanford Dogs, Oxford Flowers, and CUB datasets. Figure \ref{fig:top-10} illustrated a clear trend: as the semantic similarity decreases (from the closest adjacent class A1 to the furthest A10), retention accuracy consistently improves.

\begin{figure}[t]
  \centering
  % \fbox{\rule{0pt}{0.5in} \rule{0.9\linewidth}{0pt}}
  \includegraphics[width=1.0\linewidth]{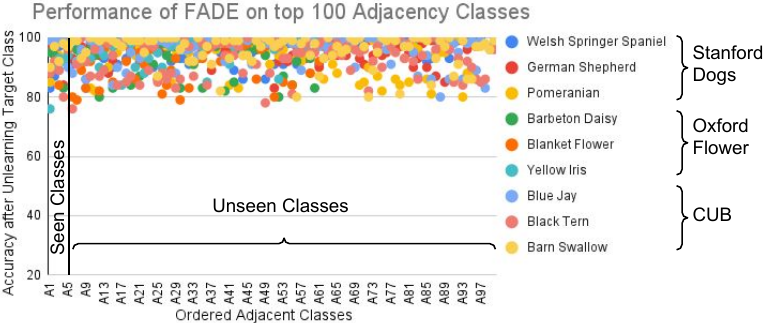}
  % \vspace{-8pt}
  \caption{Extended experiment for Adjaceny Retention from 5 unseen concepts to 95 unseen concepts. We observe that the performance remains consistent for all unseen classes. }
   % \caption{\textbf{Evaluation on concepts unseen during unlearning process} (\textcolor{blue}{cfDk}): Extended experiment from Fig. 2 (supplementary) from 5 unseen concepts to 100 unseen concepts. We observe that the performance remains consistent for all unseen classes.}
   \label{fig:unseen_exp}
    % \vspace{-18pt}
\end{figure}

To further validate this trend, we extend our analysis to the top-100 adjacent classes per target concept, where the first 5 classes are seen during training, and the remaining 95 are unseen. As shown in Figure \ref{fig:unseen_exp}, FADE consistently maintains retention accuracy above 75\% across both seen and unseen adjacent classes, demonstrating its strong generalization capability even after erasure of the target concept.

% To-Do's
% \textbf{Results of Table 1 on Top-10 classes. Detailed results of Table 2.}
% \textbf{Add reason behind choosing these concepts specifically.}
% Implementation Details.

\section{Extended Quantitative Results}
As previously discussed, we utilize Stanford Dogs, Oxford Flower, and CUB datasets to evaluate the proposed FADE and existing state-of-the-art algorithms. We present the adjacency set with their similarity scores in Table \ref{tab:similarity_scores}.

We report the classification accuracy for each class in the adjacency set of each target class from the Stanford Dogs, Oxford Flowers, and CUB datasets in Tables \ref{tab:adj_acc_stanford_dogs}, \ref{tab:adj_acc_oxford_flowers}, and \ref{tab:adj_acc_cub}. These results extend the findings reported in Table 1 of the main paper. The original model (SD) has not undergone any unlearning, so higher accuracy is better. The remaining models are comparison algorithms, and for each of them, the model should achieve lower accuracy on the target class to demonstrate better unlearning and higher accuracy on neighboring classes to show better retention of adjacent classes. From Tables \ref{tab:adj_acc_stanford_dogs}, \ref{tab:adj_acc_oxford_flowers}, and \ref{tab:adj_acc_cub}, it is evident that FADE effectively erases the target concept while preserving adjacent ones, outperforming the comparison algorithms by a significant margin across all three datasets, followed by SPM and CA. This demonstrates the superior capability at erasure and retention of the proposed FADE algorithm.

\section{Extended Qualitative Results}
% To provide a focused and detailed view of the results, Figures \ref{fig:FG_qualitative} and \ref{fig:CG_Qualitative} (borrowed from the main paper) visualize the performance of various unlearning algorithms.

\begin{figure*}[t]
\centering
  \includegraphics[width=\textwidth]{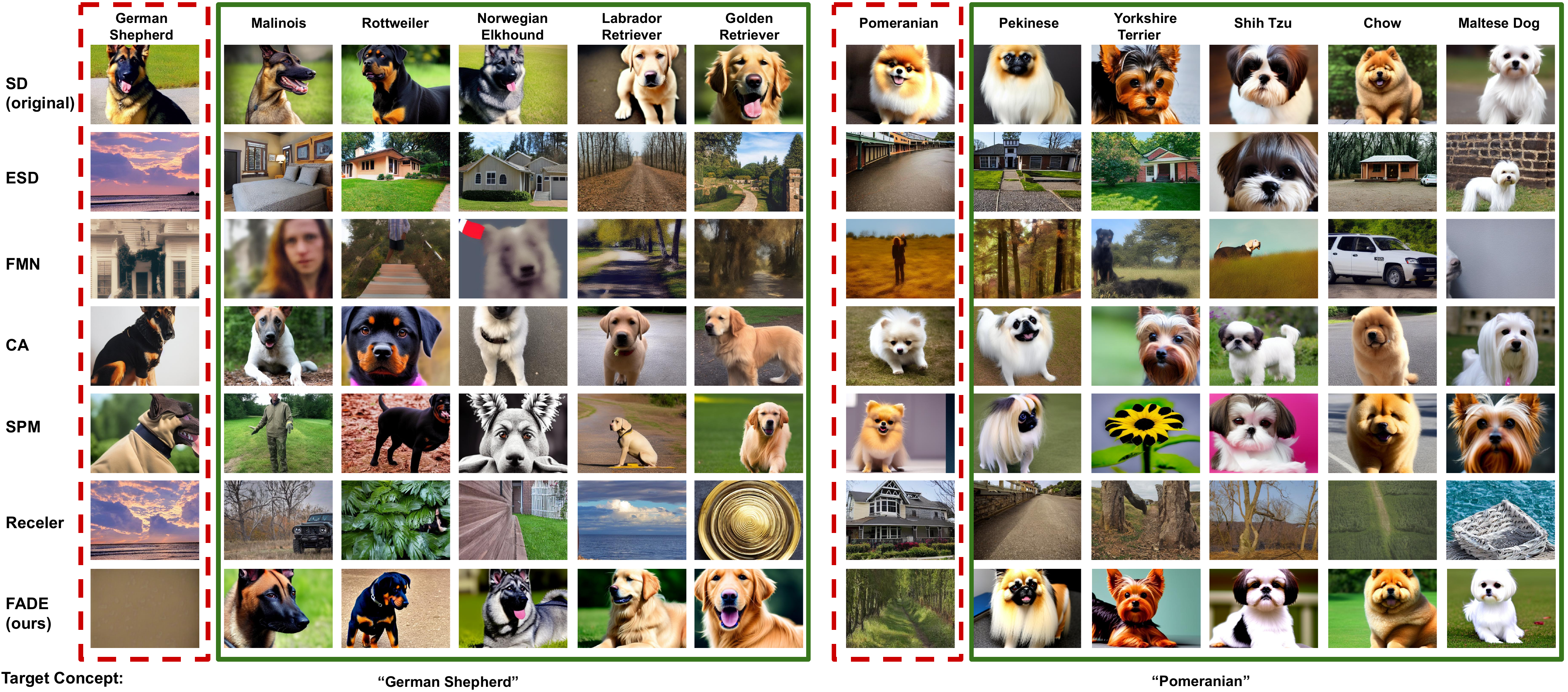}
\caption{Qualitative comparison of FADE with various algorithms for erasing German Shepherd and Pomeranian while retaining closely looking breeds extracted through concept lattice from the Stanford Dogs dataset.} 
    \label{fig:dogs_quali}
\end{figure*}

\begin{figure*}[t]
\centering
  \includegraphics[width=\textwidth]{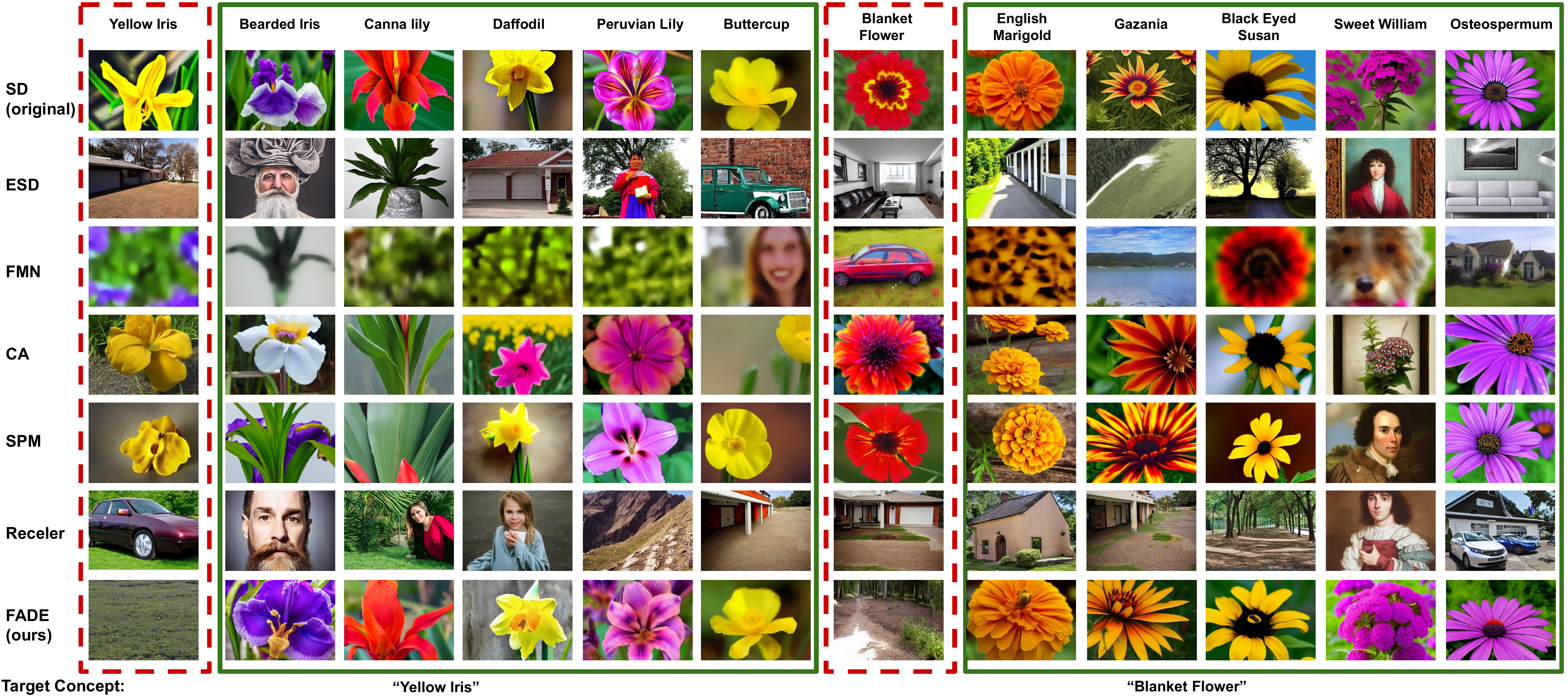}
\caption{Qualitative comparison of FADE with various algorithms for erasing Yellow Iris and Blanket Flower while retaining other similar-looking flowers through concept lattice from the Oxford Flowers dataset.} 
    \label{fig:flowers_quali}
\end{figure*}

\begin{figure*}[t]
\centering
  \includegraphics[width=\textwidth]{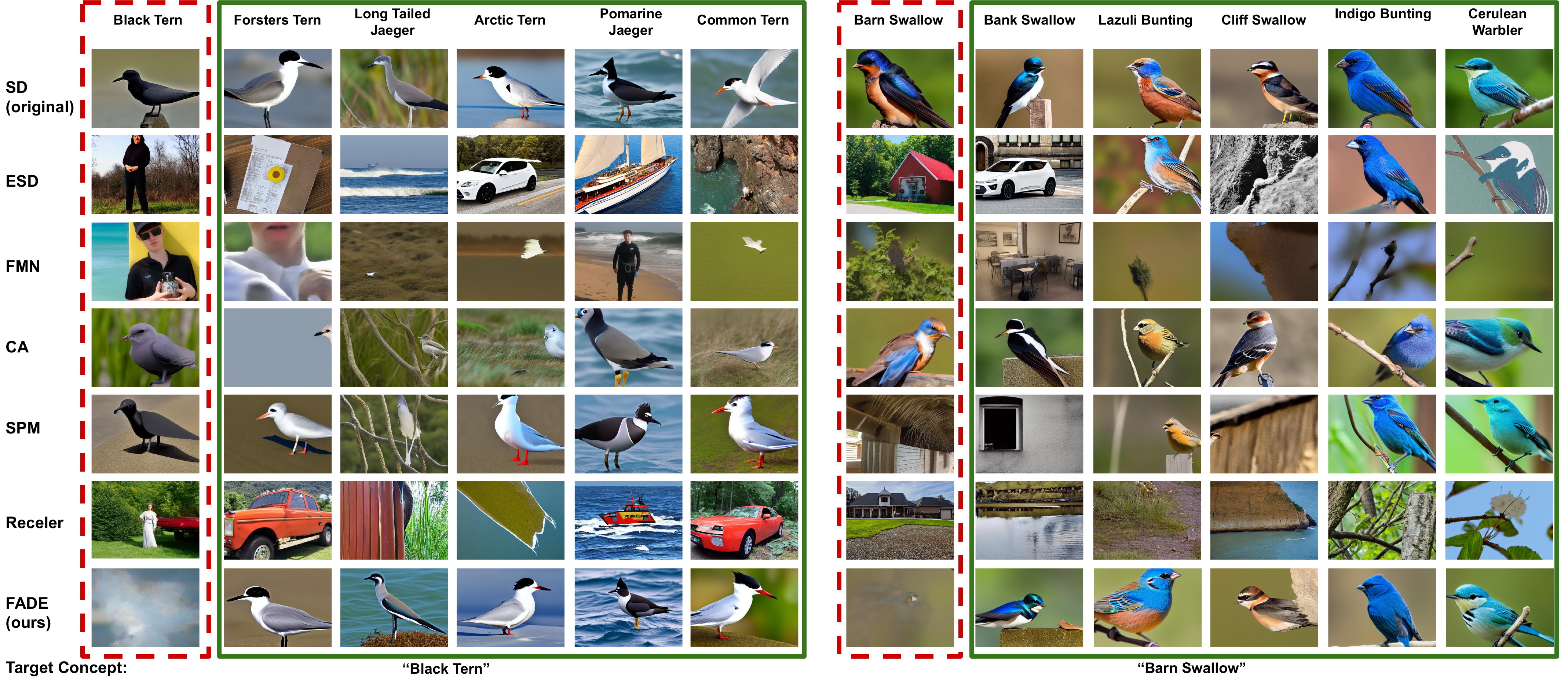}
\caption{Qualitative comparison of FADE with various algorithms for erasing Blank Tern and Barn Swallow while retaining other closely looking bird species extracted through concept lattice from CUB dataset.} 
    \label{fig:birds_quali}
\end{figure*}

Figure \ref{fig:FG_qualitative}, \ref{fig:dogs_quali}, \ref{fig:flowers_quali}, and \ref{fig:birds_quali} present the generation results for one target class and its adjacency set from each dataset, before and after applying unlearning algorithms. The first row in each of these figures displays images generated by the original Stable Diffusion (SD) model, followed by outputs from each unlearning method. Consistent with the quantitative results in Table 1 (main paper), ESD, FMN, and Receler fail to retain fine-grained details of neighboring classes. CA and SPM perform slightly better, retaining general structural features, but they struggle with specific attributes such as color and texture, especially in examples like dog breeds (e.g., Brittany Spaniel, Cocker Spaniel), bird species (e.g., Florida Jay, Cardinal), and flower species. These methods often result in incomplete erasure of the target concept or poor retention of neighboring classes.

In contrast, FADE achieves a superior balance by effectively erasing the target concept while preserving the fine-grained details of related classes, as demonstrated by the sharper distinctions in the adjacency sets. FADE’s capability is further evaluated on ImageNet-1k for target classes such as Balls, Trucks, Dogs, and Fish. Table \ref{tab:similarity_scores} lists the neghiboring classes identified using Concept Lattice to construct the adjacency set for each target class. Notably, adjacency sets generated by Concept Lattice closely align with the manually curated fine-grained class structures reported by Peychev et al. \cite{peychev2023automated}, validating the accuracy and reliability of Concept Lattice.

As shown in Table 2 of the main paper, FADE outperforms all baseline methods, achieving at least a 12\% higher ERB score compared to SPM, the next-best algorithm. FMN and CA exhibit poor performance in both adjacency retention and erasure tasks, highlighting the robustness of FADE in fine-grained unlearning scenarios.

Further, human evaluation results for FADE and baseline algorithms are presented in Table \ref{tab:user_study}, capturing erasing accuracy ($A_{\text{er}}$) and average adjacency retention accuracy ($\hat{A}_{\text{adj}}$). We also capture their balance through the proposed Erasing-Retention Balance (ERB) score.

According to human evaluators, Receler achieves the highest $A_{\text{er}}$ (86.66\%) but fails in adjacency retention, with $\hat{A}{\text{adj}}$ close to zero, resulting in a minimal ERB score (0.06). FMN and CA show suboptimal performance, with FMN favoring erasure and CA favoring retention, yielding ERB scores of 43.07 and 38.43, respectively.

FADE outperforms all baselines with the highest ERB score (59.49), balancing effective erasure ($A_{\text{er}}$ of 51.94\%) and strong adjacency retention ($\hat{A}_{\text{adj}}$ of 69.62\%). These results highlight FADE’s ability to achieve adjacency-aware unlearning without significant collateral forgetting, setting a benchmark for fine-grained erasure tasks.

\begin{table}[t]
\centering
\begin{tabular}{lccc}
\hline
            & $A_{\text{er}}$        & $\hat{A}_{\text{adj}}$      & ERB   \\ \hline
ESD         & 73.33          & 37.22          & 49.38 \\
FMN         & 49.16          & 38.33          & 43.07 \\
CA          & 30.13          & 53.05          & 38.43 \\
SPM         & 40.83          & 61.66 & 49.13 \\
Receler     & \textbf{86.66} & 0.03           & 0.06  \\
FADE (ours) & 51.94          & \textbf{69.62}         & \textbf{59.49} \\ \hline
\end{tabular}
\caption{\label{tab:user_study}\textbf{Comparison of FADE with state-of-the-art unlearning methods based on evaluations by human participants.} If prediction of human evaluator is correct, a score of 1 was given; otherwise, a score of 0 was given. The performance is reported as a percentage. According to the user study, FADE effectively balances the erasure of the target concept with the retention of neighboring concepts.}
\end{table}

\section{Implementation Details}
For all experiments and comparisons, we use Stable Diffusion v1.4 (SD v1.4) as the base model. The datasets constructed (discussed in Section 3.3 of the main paper) are generated using SD v1.4, and the same model is used to generate images for building the Concept Lattice. In Equation 9 (of the main paper), we set the base parameters as $\lambda_{\text{er}}$: 3.0,  $\lambda_{\text{adj}}$: 1000, $\lambda_{\text{guid}}$: 50. These values may vary depending on the specific target class being unlearned. For equation 6, value of $\delta$ is 1.0 across all experiments. Throughout all experiments, we optimize the model using AdamW, training for 500 iterations with a batch size of 4. All the experiments are performed on one 80 GB Nvidia A100 GPU card.

For all baseline algorithms, we utilize their official GitHub repositories and fine-tune only the cross-attention layers wherever applicable(ESD\cite{esd}, CA\cite{CA}). In the case of CA\cite{CA}, each target class is assigned its superclass as an anchor concept. For instance, for the Welsh Springer Spaniel, the anchor concept is its superclass, dog. Similarly, for concepts in the Stanford Dogs dataset, the anchor concept is set to dog, while for the Oxford Flowers dataset, it is flower, and for CUB, it is bird. This selection strategy is consistently applied when defining preservation concepts while evaluating UCE.

For calculation of $A_{\text{er}} $ and $\hat{A}_{\text{adj}}$ in Table 1 of main paper, we utilize ResNet50 as the classification model. Specifically, we fine-tune ResNet50 on 1000 images generated for each class in Stanford Dogs, Oxford Flowers and CUB datasets. For ImageNet classes in Table 2 and Table 3 of main paper  we utilize pre-trained ResNet50. For I2P related evaluations, we utilize NudeNet.

\section{Additional Analysis} 
\noindent\textbf{Choosing an adjacent concept for CA:} We conduct an additional experiment using English Springer as the anchor concept for Welsh Springer Spaniel in Concept Ablation \cite{CA}. This yields an ERB score of 69.40, significantly lower than FADE’s 95.97. While WSS$\rightarrow$ES improves erasure compared to WSS$\rightarrow$Dog, it severely degrades retention ($\hat{A}_\text{adj}$=61.4), indicating disruption in the learned manifold.

\vspace{1mm}
\noindent\textbf{Adversarial Robustness:} To assess FADE's resilience against adversarial prompts, we conducted an experiment using the Ring-a-Bell! \cite{ringabell} adversarial prompt generation algorithm. For Table 1 of the main paper, we evaluated prompts on both the original and unlearned models across Stanford Dogs, Oxford Flowers, and CUB datasets. The target class accuracies (lower is better) for the original model were 92.8, 65.4, and 45.8, while FADE significantly reduced them to 20.8, 1.3, and 5.4, demonstrating strong robustness against adversarial prompts.
% talk about CA and adversarial experiments 

\begin{figure*}[t]
\centering
  \includegraphics[width=\textwidth]{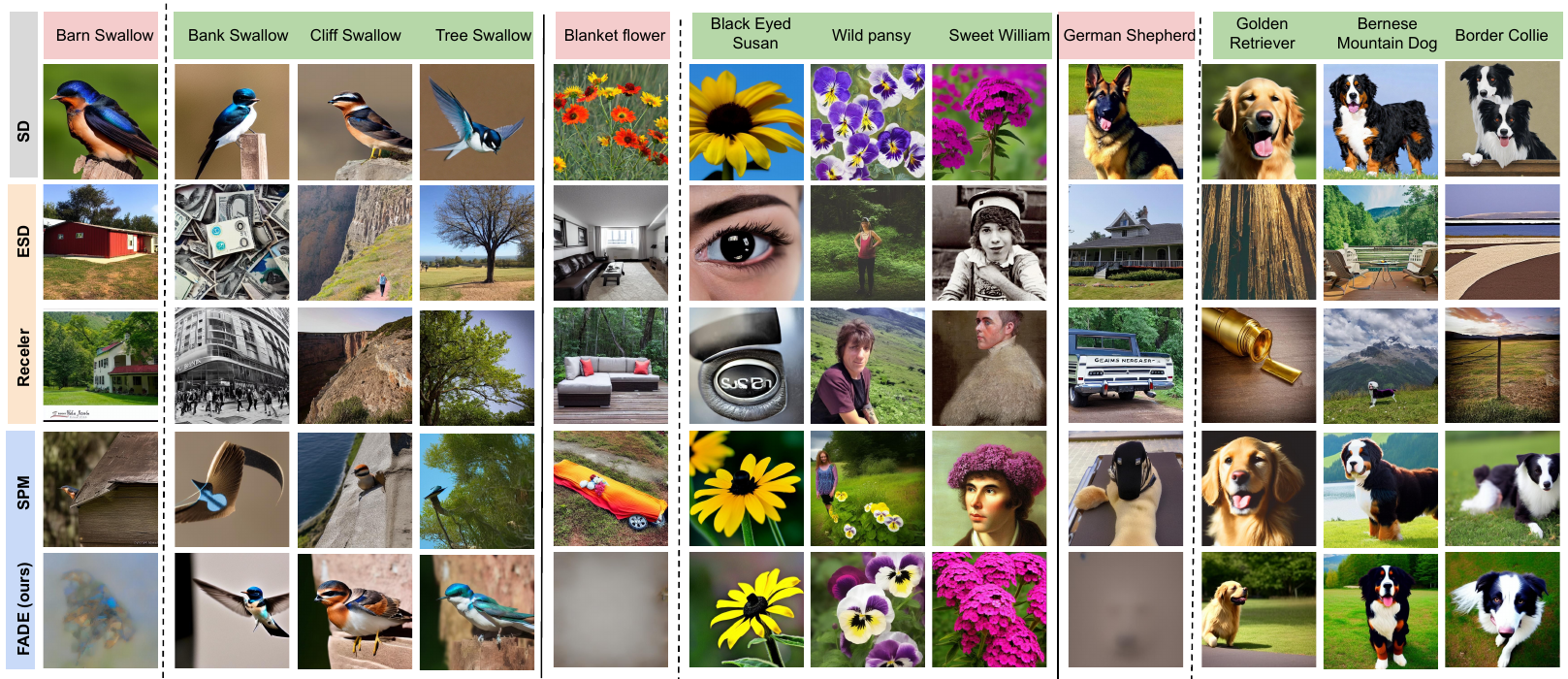}
\caption{  \label{fig:redirection}Illustration of concept redirection observed after unlearning target concepts using various algorithms. For ESD and Receler, the erasure of “Blanket Flower” redirects to unrelated outputs, such as a “girl with a black eye” for “Black-eyed Susan flower” and “a man named William” for “Sweet William flower.” Similar redirection is seen with bird classes like “Cliff Swallow” and “Tree Swallow.” In contrast, SPM and FADE effectively erase target concepts without inducing semantic redirection, ensuring coherence and retention of related knowledge.} 
  
\end{figure*}

\vspace{1mm}
\noindent\textbf{Concept Unlearning Induces Concept Redirection:} Our experiments reveal an intriguing phenomenon where unlearning a target concept often results in its redirection to an unrelated concept. As illustrated in Figure \ref{fig:redirection}, this effect is particularly evident with algorithms like ESD and Receler. For example, after unlearning the “Blanket Flower,” the model generates a “girl with a black eye” when prompted for “Black-eyed Susan flower” and produces an image of “a man named William” for the prompt “Sweet William flower.” Similarly, for bird classes such as “Cliff Swallow” and “Tree Swallow,” the unlearning process redirects the concepts to unrelated outputs, such as trees or cliffs.

Interestingly, this redirection is primarily observed in algorithms like ESD and Receler, which struggle to maintain semantic coherence post-unlearning. In contrast, SPM and the proposed FADE algorithm demonstrate robust performance, effectively erasing the target concept without inducing unintended redirections, thereby preserving the model’s semantic integrity.

% Table for Fine-Grained Unlearning on Stanford Dogs dataset
\begin{table*}[t]
\centering
\footnotesize
\begin{tabular}{llccccccc}
\hline
                                                        &                                             & SD (Original) & ESD & FMN  & CA   & SPM  & Receler & FADE (ours) \\ \hline
\multicolumn{1}{l|}{Target Concept - 1}                 & \multicolumn{1}{l|}{Welsh Springer Spaniel} & 99.10          & 0.00   & 1.24  & 37.00   & 42.27 & 0.00       & 0.45         \\ \hline
\multicolumn{1}{l|}{\multirow{5}{*}{Adjacent Concepts}} & \multicolumn{1}{l|}{Brittany Spaniel}       & 95.42          & 0.00   & 0.00    & 58.00   & 54.60 & 0.00       & 81.85        \\
\multicolumn{1}{l|}{}                                   & \multicolumn{1}{l|}{English Springer}       & 89.72          & 0.00   & 0.00    & 51.00   & 24.40 & 0.00       & 89.86        \\
\multicolumn{1}{l|}{}                                   & \multicolumn{1}{l|}{English Setter}         & 94.00            & 0.00   & 0.00    & 56.84 & 73.85 & 0.00       & 93.00          \\
\multicolumn{1}{l|}{}                                   & \multicolumn{1}{l|}{Cocker Spaniel}         & 98.85         & 40.00  & 0.00    & 82.25 & 84.10 & 0.00       & 98.82        \\
\multicolumn{1}{l|}{}                                   & \multicolumn{1}{l|}{Sussex Spaniel}         & 99.68          & 60.62  & 0.00    & 85.00   & 88.75 & 12.00      & 99.65        \\ \hline
                                                        &                                             &               &     &      &      &      &         &             \\ \hline
\multicolumn{1}{l|}{Target Concept - 2}                 & \multicolumn{1}{l|}{German Shepherd}        & 99.62          & 0.00   & 0.00    & 20.85 & 0.89  & 0.00      & 0.00           \\ \hline
\multicolumn{1}{l|}{\multirow{5}{*}{Adjacent Concepts}} & \multicolumn{1}{l|}{Malinois}               & 99.00            & 0.00   & 0.00    & 51.86 & 57.43 & 0.00       & 96.29        \\
\multicolumn{1}{l|}{}                                   & \multicolumn{1}{l|}{Rottweiler}             & 98.10          & 0.00   & 0.25  & 54.58 & 70.24 & 0.00       & 94.25        \\
\multicolumn{1}{l|}{}                                   & \multicolumn{1}{l|}{Norwegian elkhound}     & 99.76          & 10.00  & 0.00    & 63.00   & 59.00   & 0.00       & 99.65        \\
\multicolumn{1}{l|}{}                                   & \multicolumn{1}{l|}{Labrador retriever}     & 95.00            & 30.00  & 1.86  & 73.60 & 73.65 & 0.00       & 88.20        \\
\multicolumn{1}{l|}{}                                   & \multicolumn{1}{l|}{Golden retriever}       & 99.86          & 60.00  & 0.88 & 75.44 & 93.81 & 14.00      & 99.44        \\ \hline
                                                        &                                             &               &     &      &      &      &         &             \\ \hline
\multicolumn{1}{l|}{Target Concept - 3}                 & \multicolumn{1}{l|}{Pomeranian}             & 99.84          & 0.00   & 1.85  & 32.00   & 66.49 & 0.00       & 0.24        \\ \hline
\multicolumn{1}{l|}{\multirow{5}{*}{Adjacent Concepts}} & \multicolumn{1}{l|}{Pekinese}               & 98.27          & 0.00   & 0.00    & 64.63 & 86.29 & 0.00       & 84.24       \\
\multicolumn{1}{l|}{}                                   & \multicolumn{1}{l|}{Yorkshire Terrier}      & 99.90          & 40.00  & 0.00    & 89.00   & 98.62 & 0.64     & 99.62        \\
\multicolumn{1}{l|}{}                                   & \multicolumn{1}{l|}{Shih Tzu}               & 98.85          & 60.00  & 0.00    & 92.00   & 95.65 & 2.55       & 93.65        \\
\multicolumn{1}{l|}{}                                   & \multicolumn{1}{l|}{Chow}                   & 100.00           & 10.00  & 1.20  & 87.60 & 98.22 & 3.27     & 100.00         \\
\multicolumn{1}{l|}{}                                   & \multicolumn{1}{l|}{Maltese dog}            & 99.45          & 60.00  & 1.86  & 88.88 & 97.45 & 0.00       & 96.45        \\ \hline
\end{tabular}
\caption{\label{tab:adj_acc_stanford_dogs}\textbf{Comparison of Classification Accuracy on Stanford Dogs Dataset}. We compare the classification accuracy (in \%) of various models on classes from the Stanford Dogs dataset before and after unlearning on each class in the adjacency set. The original model (SD) has not undergone any unlearning (higher accuracy is better), while the rest are comparison unlearning algorithms. For each algorithm, the model should exhibit lower accuracy on the target class and higher accuracy on the adjacent concepts. It is evident that FADE significantly outperforms all the comparison algorithms.}
\end{table*}

% Table for Fine-Grained Unlearning on Oxford Flowers dataset
\begin{table*}[t]
\centering
\footnotesize
\begin{tabular}{llccccccc}
\hline
                                                        &                                        & SD (Original) & ESD & FMN  & CA    & SPM  & Receler & Ours \\ \hline
\multicolumn{1}{l|}{Target Concept}                     & \multicolumn{1}{l|}{Barbeton Daisy}    & 91.50          & 0.00   & 24.45 & 29.89  & 30.00   & 0.00       & 0.12 \\ \hline
\multicolumn{1}{l|}{\multirow{5}{*}{Adjacent Concepts}} & \multicolumn{1}{l|}{Oxeye-Daisy}       & 99.15          & 20.00  & 4.20  & 75.60  & 86.86 & 1.85     & 95.24 \\
\multicolumn{1}{l|}{}                                   & \multicolumn{1}{l|}{Black Eyed Susan}  & 97.77          & 70.00  & 2.20  & 79.08 & 94.00   & 6.00       & 94.25 \\
\multicolumn{1}{l|}{}                                   & \multicolumn{1}{l|}{Osteospermum}      & 99.50          & 50.00  & 0.85  & 90.20  & 94.80 & 0.60    & 95.65 \\
\multicolumn{1}{l|}{}                                   & \multicolumn{1}{l|}{Gazania}           & 93.50          & 0.00   & 1.00    & 62.28  & 83.84 & 0.00       & 77.45 \\
\multicolumn{1}{l|}{}                                   & \multicolumn{1}{l|}{Purple Coneflower} & 99.80          & 100.00 & 1.65  & 85.80  & 98.85 & 23.22    & 99.87 \\ \hline
                                                        &                                        &               &     &      &       &      &         &      \\ \hline
\multicolumn{1}{l|}{Target Concept}                     & \multicolumn{1}{l|}{Yellow Iris}       & 99.30          & 0.00   & 0.00    & 32.45  & 51.69 & 0.00       & 0.00    \\ \hline
\multicolumn{1}{l|}{\multirow{5}{*}{Adjacent Concepts}} & \multicolumn{1}{l|}{Bearded Iris}      & 85.25          & 0.00   & 0.00    & 20.27  & 63.60 & 0.65     & 78.68 \\
\multicolumn{1}{l|}{}                                   & \multicolumn{1}{l|}{Canna Lily}        & 98.72          & 0.00   & 0.00    & 54.45  & 76.48 & 1.63    & 95.68 \\
\multicolumn{1}{l|}{}                                   & \multicolumn{1}{l|}{Daffodil}          & 94.65          & 10.00  & 5.25  & 59.89  & 88.00   & 0.00       & 92.45 \\
\multicolumn{1}{l|}{}                                   & \multicolumn{1}{l|}{Peruvian Lily}     & 98.50          & 20.00  & 0.00    & 64.00    & 88.20 & 0.00       & 93.45 \\
\multicolumn{1}{l|}{}                                   & \multicolumn{1}{l|}{Buttercup}         & 98.00            & 0.00   & 32.00   & 78.46  & 92.23 & 0.48     & 94.00   \\ \hline
                                                        &                                        &               &     &      &       &      &         &      \\ \hline
\multicolumn{1}{l|}{Target Concept}                     & \multicolumn{1}{l|}{Blanket Flower}    & 99.50          & 0.00   & 37.00   & 73.00    & 46.00   & 0.00       & 0.00    \\ \hline
\multicolumn{1}{l|}{\multirow{5}{*}{Adjacent Concepts}} & \multicolumn{1}{l|}{English Marigold}  & 99.56          & 0.00   & 3.00    & 94.25  & 98.00   & 0.00      & 99.43 \\
\multicolumn{1}{l|}{}                                   & \multicolumn{1}{l|}{Gazania}           & 93.55          & 0.00   & 0.00    & 66.87  & 74.24 & 0.00       & 83.00   \\
\multicolumn{1}{l|}{}                                   & \multicolumn{1}{l|}{Black Eyed Susan}  & 97.77          & 0.00   & 0.47  & 72.84  & 93.45 & 1.27     & 97.00   \\
\multicolumn{1}{l|}{}                                   & \multicolumn{1}{l|}{Sweet William}     & 97.75          & 0.00   & 0.68  & 66.62  & 70.00   & 2.45     & 93.88 \\
\multicolumn{1}{l|}{}                                   & \multicolumn{1}{l|}{Osteospermum}      & 99.50          & 20.00  & 0.25  & 92.68  & 86.45 & 0.83     & 83.20 \\ \hline
\end{tabular}
\caption{\label{tab:adj_acc_oxford_flowers}\textbf{Comparison of Classification Accuracy on Oxford Flower Dataset}. We compare the classification accuracy (in \%) of various models on classes from the Oxford Flower dataset before and after unlearning on each class in the adjacency set. The original model (SD) has not undergone any unlearning (higher accuracy is better), while the rest are comparison unlearning algorithms. For each algorithm, the model should exhibit lower accuracy on the target class and higher accuracy on the adjacent concepts. It is evident that FADE significantly outperforms all the comparison algorithms.}
\end{table*}

% Table for Fine-Grained Unlearning on CUB dataset

\begin{table*}[t]
\centering
\footnotesize
\begin{tabular}{llccccccc}
\hline
                                                        &                                              & SD (Original) & ESD      & FMN  & CA   & SPM  & Receler & Ours \\ \hline
\multicolumn{1}{l|}{Target Concept}                     & \multicolumn{1}{l|}{Blue Jay}                & 99.85          & 0.00        & 0.00    & 31.42 & 14.68 & 0.00       & 0.00   \\ \hline
\multicolumn{1}{l|}{\multirow{5}{*}{Adjacent Concepts}} & \multicolumn{1}{l|}{Florida Jay}             & 98.55          & 0.00        & 1.40  & 46.25 & 69.88 & 0.00       & 98.24 \\
\multicolumn{1}{l|}{}                                   & \multicolumn{1}{l|}{White Breasted Nuthatch} & 99.00            & 5.00 & 0.00    & 72.00   & 85.08 & 0.00       & 98.44 \\
\multicolumn{1}{l|}{}                                   & \multicolumn{1}{l|}{Green Jay}               & 99.90          & 10.00       & 0.26  & 52.00   & 85.00   & 0.00       & 99.20 \\
\multicolumn{1}{l|}{}                                   & \multicolumn{1}{l|}{Cardinal}                & 100.00           & 30.00      & 0.11 & 86.85 & 96.43 & 3.48     & 100.00  \\
\multicolumn{1}{l|}{}                                   & \multicolumn{1}{l|}{Blue Winged Warbler}     & 92.97          & 20.00       & 0.54  & 49.28 & 64.24 & 0.00       & 90.65 \\ \hline
                                                        &                                              &               &          &      &      &      &         &      \\ \hline
\multicolumn{1}{l|}{Target Concept}                     & \multicolumn{1}{l|}{Black Tern}              & 86.65          & 0.00        & 4.00    & 22.45 & 13.87 & 0.00       & 0.00    \\ \hline
\multicolumn{1}{l|}{\multirow{5}{*}{Adjacent Concepts}} & \multicolumn{1}{l|}{Forsters Tern}           & 92.35          & 0.00        & 2.86  & 35.29 & 41.20 & 0.00       & 90.65 \\
\multicolumn{1}{l|}{}                                   & \multicolumn{1}{l|}{Long Tailed Jaeger}      & 97.66          & 30.00       & 9.85  & 81.08 & 90.67 & 0.66     & 90.84 \\
\multicolumn{1}{l|}{}                                   & \multicolumn{1}{l|}{Artic Tern}              & 89.50          & 0.00        & 0.45  & 22.20 & 37.85 & 0.00       & 90.26 \\
\multicolumn{1}{l|}{}                                   & \multicolumn{1}{l|}{Pomarine Jaeger}         & 88.29          & 0.00        & 0.82  & 52.80 & 63.64 & 0.00       & 80.85 \\
\multicolumn{1}{l|}{}                                   & \multicolumn{1}{l|}{Common Tern}             & 98.10          & 10.00       & 0.60  & 78.20 & 77.46 & 0.00       & 96.45 \\ \hline
                                                        &                                              &               &          &      &      &      &         &      \\ \hline
\multicolumn{1}{l|}{Target Concept}                     & \multicolumn{1}{l|}{Barn Swallow}            & 99.40          & 0.00        & 1.25  & 57.06 & 7.48  & 0.00       & 0.45  \\ \hline
\multicolumn{1}{l|}{\multirow{5}{*}{Adjacent Concepts}} & \multicolumn{1}{l|}{Bank Swallow}            & 9.79           & 0.00        & 0.65  & 54.60 & 30.21 & 0.00       & 93.60 \\
\multicolumn{1}{l|}{}                                   & \multicolumn{1}{l|}{Lazuli Bunting}          & 99.75          & 70.00       & 0.65  & 82.00   & 88.40 & 0.00       & 99.00   \\
\multicolumn{1}{l|}{}                                   & \multicolumn{1}{l|}{Cliff Swallow}           & 93.20          & 0.00        & 0.00    & 77.00   & 47.63 & 0.86     & 91.25 \\
\multicolumn{1}{l|}{}                                   & \multicolumn{1}{l|}{Indigo Bunting}          & 96.80          & 70.00       & 17.46 & 87.68 & 93.00   & 5.22     & 96.45 \\
\multicolumn{1}{l|}{}                                   & \multicolumn{1}{l|}{Cerulean Warbler}        & 96.90          & 50.00       & 2.65  & 88.40 & 89.00   & 0.45     & 96.80 \\ \hline
\end{tabular}
\caption{\label{tab:adj_acc_cub}\textbf{Comparison of Classification Accuracy on CUB Dataset}. We compare the classification accuracy (in \%) of various models on classes from the CUB dataset before and after unlearning on each class in the adjacency set. The original model (SD) has not undergone any unlearning (higher accuracy is better), while the rest are comparison unlearning algorithms. For each algorithm, the model should exhibit lower accuracy on the target class and higher accuracy on the adjacent concepts. It is evident that FADE significantly outperforms all the comparison algorithms.}
\end{table*}

% Table of Adjacency Set with their similarity scores.
\begin{sidewaystable*}[]
\centering
\resizebox{\textwidth}{!}{
\begin{tabular}{lcccccc}
\hline
\multicolumn{1}{l|}{\multirow{2}{*}{\begin{tabular}[c]{@{}l@{}}Stanford Dogs\\ Dataset\end{tabular}}} & \multicolumn{2}{c|}{Welsh Springer Spaniel}                         & \multicolumn{2}{c|}{German Shepherd}                              & \multicolumn{2}{c}{Pomeranian}                 \\
\multicolumn{1}{l|}{}                                                                                 & Class Names                 & \multicolumn{1}{c|}{Similarity Score} & Class Names               & \multicolumn{1}{c|}{Similarity Score} & Class Names                 & Similarity Score \\ \hline
\multicolumn{1}{l|}{Adjacent Class - 1}                                                               & Brittany Spaniel            & \multicolumn{1}{c|}{98.19}            & Malinois                  & \multicolumn{1}{c|}{95.84}            & Pekinese                    & 92.67            \\
\multicolumn{1}{l|}{Adjacent Class - 2}                                                               & English Springer            & \multicolumn{1}{c|}{97.05}            & Rottweiler                & \multicolumn{1}{c|}{92.52}            & Yorkshire Terrier           & 92.42            \\
\multicolumn{1}{l|}{Adjacent Class - 3}                                                               & English Setter              & \multicolumn{1}{c|}{95.12}            & Norwegian Walkhound       & \multicolumn{1}{c|}{92.51}            & Shih Tzu                    & 91.58            \\
\multicolumn{1}{l|}{Adjacent Class - 4}                                                               & Cocker Spaniel              & \multicolumn{1}{c|}{95.10}            & Labrador Retriever        & \multicolumn{1}{c|}{91.63}            & Chow                        & 90.99            \\
\multicolumn{1}{l|}{Adjacent Class - 5}                                                               & Sussex Spaniel              & \multicolumn{1}{c|}{93.62}            & Golden Retriever          & \multicolumn{1}{c|}{91.43}            & Maltese Dog                 & 90.96            \\
\multicolumn{1}{l|}{Adjacent Class - 6}                                                               & Blenheim Spaniel            & \multicolumn{1}{c|}{93.05}            & Collie                    & \multicolumn{1}{c|}{90.79}            & Chihuaha                    & 90.49            \\
\multicolumn{1}{l|}{Adjacent Class - 7}                                                               & Irish Setter                & \multicolumn{1}{c|}{92.90}            & Doberman                  & \multicolumn{1}{c|}{90.37}            & Papillon                    & 90.38            \\
\multicolumn{1}{l|}{Adjacent Class - 8}                                                               & Saluki                      & \multicolumn{1}{c|}{92.83}            & Black and Tan Coonhound   & \multicolumn{1}{c|}{90.27}            & Samoyed                     & 89.22            \\
\multicolumn{1}{l|}{Adjacent Class - 9}                                                               & English Foxhound            & \multicolumn{1}{c|}{92.81}            & Bernese Mountain dog      & \multicolumn{1}{c|}{90.06}            & Australian Terrier          & 89.15            \\
\multicolumn{1}{l|}{Adjacent Class - 10}                                                              & Gordon Setter               & \multicolumn{1}{c|}{92.35}            & Border collie             & \multicolumn{1}{c|}{89.56}            & Toy poodle                  & 89.05            \\ \hline
                                                                                                      &                             &                                       &                           &                                       &                             &                  \\ \hline
\multicolumn{1}{l|}{\multirow{2}{*}{\begin{tabular}[c]{@{}l@{}}Oxford Flower\\ Dataset\end{tabular}}} & \multicolumn{2}{c|}{Barbeton Daisy}                                 & \multicolumn{2}{c|}{Yellow Iris}                                  & \multicolumn{2}{c}{Blanket Flower}             \\
\multicolumn{1}{l|}{}                                                                                 & Class Names                 & \multicolumn{1}{c|}{Similarity Score} & Class Names               & \multicolumn{1}{c|}{Similarity Score} & Class Names                 & Similarity Score \\ \hline
\multicolumn{1}{l|}{Adjacent Class - 1}                                                               & Oxeye Daisy                 & \multicolumn{1}{c|}{98.65}            & Bearded Iris              & \multicolumn{1}{c|}{96.38}            & English Marigold            & 95.13            \\
\multicolumn{1}{l|}{Adjacent Class - 2}                                                               & Black eyed susan            & \multicolumn{1}{c|}{95.56}            & Canna Lily                & \multicolumn{1}{c|}{92.48}            & Gazania                     & 92.38            \\
\multicolumn{1}{l|}{Adjacent Class - 3}                                                               & Osteospermum                & \multicolumn{1}{c|}{94.99}            & Daffodil                  & \multicolumn{1}{c|}{92.33}            & Black Eyed Susan            & 90.64            \\
\multicolumn{1}{l|}{Adjacent Class - 4}                                                               & Gazania                     & \multicolumn{1}{c|}{93.91}            & Peruvian Lily             & \multicolumn{1}{c|}{92.18}            & Sweet William               & 90.00            \\
\multicolumn{1}{l|}{Adjacent Class - 5}                                                               & Purple Coneflower           & \multicolumn{1}{c|}{93.27}            & Buttercup                 & \multicolumn{1}{c|}{91.55}            & Osteospermum                & 89.94            \\
\multicolumn{1}{l|}{Adjacent Class - 6}                                                               & Pink yellow dahlia          & \multicolumn{1}{c|}{91.74}            & Hippeastrum               & \multicolumn{1}{c|}{91.33}            & Barbeton Daisy              & 89.86            \\
\multicolumn{1}{l|}{Adjacent Class - 7}                                                               & Sunflower                   & \multicolumn{1}{c|}{91.18}            & Moon Orchid               & \multicolumn{1}{c|}{91.08}            & Purple Coneflower           & 89.45            \\
\multicolumn{1}{l|}{Adjacent Class - 8}                                                               & Buttercup                   & \multicolumn{1}{c|}{91.17}            & Ruby Lipped Cattleya      & \multicolumn{1}{c|}{90.75}            & Snapdragon                  & 89.42            \\
\multicolumn{1}{l|}{Adjacent Class - 9}                                                               & Japanese Anemone            & \multicolumn{1}{c|}{90.56}            & Hard-Leaved Pocket Orchid & \multicolumn{1}{c|}{90.36}            & Wild Pansy                  & 88.06            \\
\multicolumn{1}{l|}{Adjacent Class - 10}                                                              & Magnolia                    & \multicolumn{1}{c|}{90.27}            & Azalea                    & \multicolumn{1}{c|}{90.02}            & Pink Yellow Dahlia          & 88.01            \\ \hline
                                                                                                      &                             &                                       &                           &                                       &                             &                  \\ \hline
\multicolumn{1}{l|}{\multirow{2}{*}{CUB Dataset}}                                                     & \multicolumn{2}{c|}{Blue Jay}                                       & \multicolumn{2}{c|}{Black Tern}                                   & \multicolumn{2}{c}{Barn Swallow}               \\
\multicolumn{1}{l|}{}                                                                                 & Class Names                 & \multicolumn{1}{c|}{Similarity Score} & Class Names               & \multicolumn{1}{c|}{Similarity Score} & Class Names                 & Similarity Score \\ \hline
\multicolumn{1}{l|}{Adjacent Class - 1}                                                               & Florida Jay                 & \multicolumn{1}{c|}{93.62}            & Forsters Tern             & \multicolumn{1}{c|}{96.19}            & Bank Swallow                & 95.91            \\
\multicolumn{1}{l|}{Adjacent Class - 2}                                                               & White Breasted Nuthatch     & \multicolumn{1}{c|}{92.50}            & Long Tailed Jaeger        & \multicolumn{1}{c|}{95.54}            & Lazuli Bunting              & 93.91            \\
\multicolumn{1}{l|}{Adjacent Class - 3}                                                               & Green Jay                   & \multicolumn{1}{c|}{91.91}            & Artic Tern                & \multicolumn{1}{c|}{94.79}            & Cliff Swallow               & 92.47            \\
\multicolumn{1}{l|}{Adjacent Class - 4}                                                               & Cardinal                    & \multicolumn{1}{c|}{90.86}            & Pomarine Jaeger           & \multicolumn{1}{c|}{94.52}            & Indigo Bunting              & 92.04            \\
\multicolumn{1}{l|}{Adjacent Class - 5}                                                               & Blue Winged Warbler         & \multicolumn{1}{c|}{90.00}            & Common Tern               & \multicolumn{1}{c|}{93.35}            & Cerulean Warbler            & 91.65            \\
\multicolumn{1}{l|}{Adjacent Class - 6}                                                               & Downy Woodpecker            & \multicolumn{1}{c|}{88.84}            & Elegant Tern              & \multicolumn{1}{c|}{93.03}            & Blue Grosbeak               & 91.58            \\
\multicolumn{1}{l|}{Adjacent Class - 7}                                                               & Indigo Bunting              & \multicolumn{1}{c|}{88.83}            & Frigatebird               & \multicolumn{1}{c|}{91.32}            & Tree Swallow                & 91.33            \\
\multicolumn{1}{l|}{Adjacent Class - 8}                                                               & Cerulean Warbler            & \multicolumn{1}{c|}{88.80}            & Least tern                & \multicolumn{1}{c|}{91.28}            & Black Throated Blue Warbler & 90.43            \\
\multicolumn{1}{l|}{Adjacent Class - 9}                                                               & Black Throated Blue Warbler & \multicolumn{1}{c|}{88.64}            & Red legged Kittiwake      & \multicolumn{1}{c|}{91.22}            & Blue winged warbler         & 89.25            \\
\multicolumn{1}{l|}{Adjacent Class - 10}                                                              & Clark Nutcracker            & \multicolumn{1}{c|}{88.39}            & Lysan Albatross           & \multicolumn{1}{c|}{90.17}            & White breasted kingfisher   & 89.16            \\ \hline
\end{tabular}
}
\caption{\label{tab:similarity_scores}Description of the adjacency set for target classes from the Stanford Dogs, Oxford Flowers, and CUB datasets, along with their similarity scores.}
\end{sidewaystable*}

% \begin{sidewaysfigure}
% \begin{figure*}
% \centering
% % \includegraphics[width=\textheight]{Images/FADE_qualitative_results_1.pdf}
% \includegraphics[width=\textwidth]{Images/FADE_qualitative_results_1.pdf}
% \caption{\textbf{Qualitative comparison between existing and proposed algorithms for erasing target concepts and testing retention on neighboring fine-grained concepts.} Results are shown for one target concept each from the Stanford Dogs, Oxford Flowers, and CUB datasets.}
% \label{fig:FG_qualitative}
% \end{figure*}
% \end{sidewaysfigure}

% \begin{figure*}[t]
% \centering
%   \includegraphics[width=\textwidth]{Images/FADE_qualitative_results_1_I10.pdf}
% \caption{Comparison of FADE with various algorithms for erasing the 'garbage truck' class in Fine-Grained and Coarse-Grained Unlearning. The target class, adjacency set and the retain set and constructed from the ImageNet-1k dataset.} 
%     \label{fig:CG_Qualitative}
% \end{figure*}

\end{document}